\documentclass{article}
\usepackage[accepted]{icml2019}
\usepackage[utf8]{inputenc}
\usepackage{soul}
\usepackage[small]{caption}

\usepackage{url}            
\usepackage{booktabs}       
\usepackage{amsfonts}       
\usepackage{nicefrac}       
\usepackage{microtype}      

\usepackage{booktabs} 
\usepackage[titletoc]{appendix}

\usepackage{adjustbox}
\usepackage{blindtext}
\usepackage{times}
\usepackage{amsmath, amsthm, amssymb}
\usepackage{romannum}
\usepackage{float}
\usepackage{framed}
\usepackage{mathtools}
\usepackage{thmtools,thm-restate}
\usepackage{xspace}
\usepackage{graphicx}
\usepackage{subfigure}
\usepackage[pdf]{pstricks}
\usepackage{multirow}
\usepackage{lipsum}

\usepackage{pifont}
\newtheorem{theorem}{Theorem}[section]
\newtheorem{lemma}[theorem]{Lemma}

\newtheorem{corollary}[theorem]{Corollary}

\newenvironment{definition}[1][Definition]{\begin{trivlist}
\item[\hskip \labelsep {\bfseries #1}]}{\end{trivlist}}

\newcommand\defeq{\stackrel{\mathclap{\normalfont\mbox{\tiny{def}}}}{=}}

\DeclarePairedDelimiter\ceil{\lceil}{\rceil}
\DeclarePairedDelimiter\bceil{\Big\lceil}{\Big\rceil}
\DeclarePairedDelimiter\bbceil{\Bigg\lceil}{\Bigg\rceil}

\newcommand{\A}{\mathcal{A}}
\newcommand{\B}{\mathcal{B}}
\newcommand{\D}{\mathcal{D}}

\newcommand{\I}{\mathcal{I}}
\newcommand{\K}{\mathcal{K}}

\DeclareMathOperator{\R}{\mathop{\mathbb{R}}}
\DeclareMathOperator*{\E}{\mathop{\mathbb{E}}}
\DeclareMathOperator*{\Lim}{Lt}

\newcommand*{\wrt}{with respect to\@\xspace}
\newcommand*{\iid}{i.i.d.\@\xspace}

\makeatletter
\newcommand*{\etc}{%
    \@ifnextchar{.}%
        {etc}%
        {etc.\@\xspace}%
}
\makeatother
\makeatletter
\newcommand*{\etal}{%
    \@ifnextchar{.}%
        {{\textit{et al}}}%
        {{\textit{et al}}.\@\xspace}%
}
\makeatother

\makeatletter
\newcommand*{\UCBM}{%
    \@ifnextchar{.}%
        {{\textsc{UCB-M}}}%
    \@ifnextchar{,}%
        {{\textsc{UCB-M}}}%
    \@ifnextchar{:}%
        {{\textsc{UCB-M}}}%
        {{\textsc{UCB-M}}\@\xspace}%
}
\makeatother

\makeatletter
\newcommand*{\CRM}{%
    \@ifnextchar{.}%
        {{\textsc{CR-M}}}%
    \@ifnextchar{,}%
        {{\textsc{CR-M}}}%
    \@ifnextchar{:}%
        {{\textsc{CR-M}}}%
        {{\textsc{CR-M}}\@\xspace}%
}
\makeatother

\makeatletter
\newcommand*{\QRMUCBM}{%
    \@ifnextchar{.}%
        {{\textsc{QUCB-M}}}%
    \@ifnextchar{,}%
        {{\textsc{QUCB-M}}}%
    \@ifnextchar{:}%
        {{\textsc{QUCB-M}}}%
        {{\textsc{QUCB-M}}\@\xspace}%
}
\makeatother

\makeatletter
\newcommand*{\QRMMSM}{%
    \@ifnextchar{.}%
        {{\textsc{QMoss-M}}}%
    \@ifnextchar{,}%
        {{\textsc{QMoss-M}}}%
    \@ifnextchar{:}%
        {{\textsc{QMoss-M}}}%
    {{\textsc{QMoss-M}}\@\xspace}%
}
\makeatother

\makeatletter
\newcommand*{\QRMTSM}{%
    \@ifnextchar{.}%
        {{\textsc{QTS-M}}}%
    \@ifnextchar{,}%
        {{\textsc{QTS-M}}}%
    \@ifnextchar{:}%
        {{\textsc{QTS-M}}}%
    {{\textsc{QTS-M}}\@\xspace}%
}
\makeatother

\makeatletter
\newcommand*{\QURM}{%
    \@ifnextchar{.}%
        {{\textsc{QR-M}}}%
    \@ifnextchar{,}%
        {{\textsc{QR-M}}}%
    \@ifnextchar{:}%
        {{\textsc{QR-M}}}%
    {{\textsc{QR-M}}\@\xspace}%
}
\makeatother

\makeatletter
\newcommand*{\QRMTWO}{%
    \@ifnextchar{.}%
        {{\textsc{QRM2}}}%
    \@ifnextchar{,}%
        {{\textsc{QRM2}}}%
    \@ifnextchar{:}%
        {{\textsc{QRM2}}}%
    {{\textsc{QRM2}}\@\xspace}%
}
\makeatother

\makeatletter
\newcommand*{\UCBMPA}{%
    \@ifnextchar{.}%
        {{\textsc{UCB-MPA}}}%
    \@ifnextchar{,}%
        {{\textsc{UCB-MPA}}}%
    \@ifnextchar{:}%
        {{\textsc{UCB-MPA}}}%
        {{\textsc{UCB-MPA}}\@\xspace}%
}
\makeatother

\makeatletter
\newcommand*{\UCB}{%
    \@ifnextchar{.}%
        {{\textsc{UCB1}}}%
    \@ifnextchar{,}%
        {{\textsc{UCB1}}}%
    \@ifnextchar{:}%
        {{\textsc{UCB1}}}%
        {{\textsc{UCB1}}\@\xspace}%
}
\makeatother

\makeatletter
\newcommand*{\Moss}{%
    \@ifnextchar{.}%
        {{\textsc{Moss}}}%
    \@ifnextchar{,}%
        {{\textsc{Moss}}}%
    \@ifnextchar{:}%
        {{\textsc{Moss}}}%
        {{\textsc{Moss}}\@\xspace}%
}
\makeatother

\makeatletter
\newcommand*{\TS}{%
    \@ifnextchar{.}%
        {{\textsc{Thompson}  Sampling}}%
    \@ifnextchar{,}%
        {{\textsc{Thompson}  Sampling}}%
    \@ifnextchar{:}%
        {{\textsc{Thompson}  Sampling}}%
        {{\textsc{Thompson}  Sampling}\@\xspace}%
}
\makeatother

\makeatletter
\newcommand*{\TSM}{%
    \@ifnextchar{.}%
        {{\textsc{TS-M}}}%
    \@ifnextchar{,}%
        {{\textsc{TS-M}}}%
    \@ifnextchar{:}%
        {{\textsc{TS-M}}}%
        {{\textsc{TS-M}}\@\xspace}%
}
\makeatother

\makeatletter
\newcommand*{\MSM}{%
    \@ifnextchar{.}%
        {{\textsc{Moss-M}}}%
    \@ifnextchar{,}%
        {{\textsc{Moss-M}}}%
    \@ifnextchar{:}%
        {{\textsc{Moss-M}}}%
        {{\textsc{Moss-M}}\@\xspace}%
}
\makeatother

\makeatletter
\newcommand{\ALOOP}[1]{\ALC@it\algorithmicloop\ #1%
  \begin{ALC@loop}}
\newcommand{\ENDALOOP}{\end{ALC@loop}\ALC@it\algorithmicendloop}
\renewcommand{\algorithmicrequire}{\textbf{Input:}}

\makeatother

\begin{document}
\twocolumn[
\icmltitle{Regret Minimisation in Multi-Armed Bandits Using Bounded Arm Memory}
\icmlsetsymbol{equal}{*}
\begin{icmlauthorlist}
\icmlauthor{Arghya Roy Chaudhuri}{iitb}
\icmlauthor{Shivaram Kalyanakrishnan}{iitb}
\end{icmlauthorlist}

\icmlaffiliation{iitb}{Department of Computer Science and Engineering, Indian Institute of Technology Bombay, Mumbai 400076, India}

\icmlcorrespondingauthor{Arghya Roy Chaudhuri}{arghya@cse.iitb.ac.in}
\icmlcorrespondingauthor{Shivaram Kalyanakrishnan}{shivaram@cse.iitb.ac.in}

]
\printAffiliationsAndNotice{}
\begin{abstract}
In this paper, we propose a constant word (RAM model) algorithm for regret minimisation
for both finite and infinite Stochastic Multi-Armed Bandit (MAB) instances.
Most of the existing regret minimisation algorithms
need to remember the statistics of all the arms  they encounter. This may become a problem
for the cases where the number of available words of memory is limited.

Designing an efficient regret minimisation algorithm that uses a constant number of words has long been
interesting to
the community. Some early attempts consider the number of arms to be infinite, and
require the reward distribution of the arms to belong to some particular family.
Recently, for finitely many-armed bandits an explore-then-commit based 
algorithm~\citep{Liau+PSY:2018} seems to escape such assumption. However, due to
the underlying PAC-based elimination their method incurs a high regret. We present a conceptually simple,
and efficient algorithm that needs to remember statistics of at most $M$ arms, and
for any $K$-armed finite bandit instance it
enjoys a $O(KM +K^{1.5}\sqrt{T\log (T/MK)}/M)$ 
 upper-bound on regret.
We extend it to achieve sub-linear \textit{quantile-regret}~\citep{RoyChaudhuri+K:2018} and empirically verify the efficiency of our algorithm via experiments. 
\end{abstract}
\section{Introduction}
In this paper, we investigate the problem of regret minimisation in
Multi-Armed Bandit (MAB)~\citep{Berry+F:1985} using a bounded number of words.
Each \emph{arm} in a bandit instance represents a slot-machine with a
fixed (but unknown) real-valued reward distribution associated with it.
At each time step, the experimenter is supposed to select and
\emph{pull} an arm, and observe the reward. The goal of the experimenter
is to maximise the expected total reward for a finite time horizon, thereby
minimising the expected \emph{regret} measured with respect to the mean of the
optimal arm.


A range of real-world applications like drug testing~\citep{Armitage:1960,Colton:1963},
crowd-sourcing~\citep{Tran-Thanh+SRJ:2014} \etc
can be modelled using multi-armed bandits, where the number of arms is high.
In such cases, due to budgetary constraints or some other practical considerations, 
it might viable to experiment only with a small number of arms instead of the
whole pool. The problem is of particular interest because the experimenter gets
to store statistics of a small but fixed number of arms. Therefore, it adds another
layer of an exploration-exploitation dilemma for the task of regret minimisation.
This particular set-up has been drawing attention since long ago~\cite{Cover:1968}; however,
only a few investigations have been made in this direction.

In this paper, we present a regret minimisation algorithm that uses a bounded number of words,
for both finite and infinite-armed bandits. Unlike the existing algorithms, our
algorithm does not need any special assumption of reward distribution of arms but bypass the explicit PAC-based exploration for the sake of efficiency.
Below, we formalise our problem followed by our specific contributions.

\paragraph{Background and Problem Setup.} A bandit-instance $\B = (\A, \D)$ consists
of a set of arms $\A$, and a set of sub-Gaussian cumulative distribution
functions (CDF) $\D$.
Each arm $a \in \A$, when pulled, generates a \iid reward from the 
corresponding CDF $D_a \in \D$, defined over $[0,1]$.
The expected reward of arm $a\in \A$ is given by $\mu_a \defeq \E_{r \sim D_a}[r]$. 
We also assume that the experimenter has no information regarding $\D$.
The only way for her to gather knowledge about $\D$ is via generated rewards by 
sampling the arms. 
We define a set called history as $H_t \defeq \{(a_i, r_i)\}_{i=1}^{t}$,
where, $r_i\in [0, 1]$ is the reward produced at $i$-th step by pulling the arm $a_i \in \A$.

\paragraph{Cumulative Regret Minimisation.} 
Assuming
$\mu^* \defeq \min\{y \in [0,1]: \forall a \in \A, \mu(a) \leq y\}$, and the 
given horizon of pulls as $T$, the conventional cumulative regret incurred by a 
algorithm is defined as

\begin{equation}
 \label{eq:cumregdef}
 \R_T^* \defeq T\mu^* - \sum_t^T \E[\mu_{a_t}],
\end{equation}

wherein $a_t$ is the arm pulled by the algorithm at time $t$. The expectation is
taken over random rewards and the possible randomisation introduced by the algorithm.

We briefly restate the definition of ``quantile regret''
introduced by~\cite{RoyChaudhuri+K:2018} based on their previous contribution in
a pure exploration setting~\citep{RoyChaudhuri+K:2017}. A problem instance $\I = (\B, P_\A)$
consists of a bandit instance $\B$, and a sampling distribution for choosing arm
from $\A$. Letting $\rho \in [0,1]$, the $(1-\rho)$-th quantile of $P_\A$ is
defined as
\begin{equation}
 \mu_\rho = \inf\{x \in [0,1]: \Pr_{a\sim P_\A}\{\mu_a \leq x\}\geq 1-\rho\}.
\end{equation}

Then, for a given horizon of pulls as $T$, quantile regret with respect to
$\mu_\rho$ is defined as 
\begin{equation}
 \label{eq:cumregdef}
 \R_T(\rho) \defeq T\mu_\rho - \sum_t^T \E[\mu_{a_t}],
\end{equation}
wherein, $a_t$ and $\E[\cdot]$ bear the same interpretation.

\paragraph{RAM Model.} 
It should be noted that given any bandit instance $\B = (\A, \D)$,
as we are not considering any special structure in
$\A$ or $\D$, putting a restriction on an algorithm to use a bounded number of words of space,
either restricts the horizon of pulls,
or restricts the algorithm to store statistics of
only bounded number of arms simultaneously. In this paper, we consider the latter and assume $M$ to be
that number.
We adopt the word RAM model~\citep{Aho+HU:1974,Cormen+LRS:2009},
that considers a word as the unit of space.
This model facilitates to consider that each of the input values and variables can be
stored in $O(1)$ word space. For finite bandit instances ($|\A| < \infty$),
we consider a word to be consisted of $O(\log T)$ bits. Therefore, our algorithm needs space-complexity of
$O(M \log T + \log |\A|)$ bits. For the infinite 
bandit instances ($|\A| = \infty$), for $\rho \in [0,1]$, if the experimenter
needs to analyse the performance with respect to $\mu_\rho$ (the $(1-\rho)$-th quantile),
she must allow the algorithm to use $O(M\log T + \log (1/\rho))$ bits.

We call this set of arm indices whose statistics are stored
as \emph{arm memory} and its cardinality as \emph{arm memory size}. Hence,
an algorithm with arm memory size $M$ can store the statistics of at most $M$
arms. Also, it should be noted that an algorithm is allowed to pull an arm only
if it is stored in the memory. Hence, before pulling a new arm
(which is not currently in the arm memory), the algorithm should replace
an arm in its arm memory with this new arm. It is interesting to note that
the algorithms that work with $M=1$,
can only keep the stat of the arm it is currently pulling. 
Therefore, switching to a new arm costs such an algorithm to lose all the
experience gained by sampling the previous arm. However, for a finite bandit instance,
as the algorithms are allowed to remember all the arm indices, such an algorithm can
store the gained experience by storing a bounded number of arm indices for possible further special
treatment. The scenario is widely different for infinite bandit instances, where an
algorithm can pull a new arm only if it is chosen by the given sampling distribution
$P_\A$. In such a scenario, once an algorithm discards an arm from the arm memory,
it can encounter that arm only if it is sampled again in future by $P_\A$. Hence, the
algorithm can not recall a discarded arm. In the existing literature~\citep{Herschkorn+PR:1996, Berry+CZHS:1997} 
on infinite bandit instances, such algorithms are termed as \emph{non-recalling} algorithms.
However, for $M > 1$ (but bounded above), an algorithm enjoys the freedom of ensuring a 
previously encountered good arm to keep in the memory, irrespective of whether or not the
bandit instance is finite or not. Our findings show that this is more beneficial than
the non-recalling algorithms for infinite bandit instances.

 
 \paragraph{Problem Definition.} Given a positive integer $M$, below, we define the problem of  \underline{c}onventional \underline{r}egret \underline{m}inimisation
 (\textsc{CR-M}) 
 and extend the definition to 
 \underline{q}uantile \underline{r}egret \underline{m}inimisation
 (\textsc{QR-M}).
 
\begin{definition}[\protect\CRM.]
 An algorithm $\mathcal{L}$ is said to \emph{solve} \CRM, if takes $\A$, and $M$
 as the input and for a sufficiently large budget $T$ (not necessarily known beforehand)
 it will achieve $\R_T^* \in o(T)$; using an \emph{arm memory size} at most $M$.
 It is assumed that for a finite bandit instance with $|\A| = K < \infty$, 
 the algorithm is allowed to store $O(M\log(T) + \log K)$ bits of information.
\end{definition}

\begin{definition}[\protect\QURM.]
 Suppose we are given a problem instance $\I = (\B, P_\A)$, and a positive integer
 $M$. Let, $\rho_0 \in (0,1]$.
 An algorithm $\mathcal{L}$ is said to \emph{solve} \QURM, if takes $\A$, $P_\A$, and $M$
 as the input, and for
 every $\rho \in (\rho_0, 1]$, given a sufficiently large budget $T$
 it will achieve $\R_T(\rho) \in o(T)$; using an \emph{arm memory size} at most $M$.
 It is assumed that the algorithm is allowed to store  $O(M\log(T) + \log (1/\rho_0))$ bits of information.
\end{definition}

In this paper we present algorithms solve \CRM on finite bandit instances, and \QURM,
with $M \geq 2$. 
Following we brief our contribution in this paper.

\paragraph{Contributions.} We present algorithms for minimisation of conventional regret and quantile regret, using a bounded number of words of space. Following
is the list of our specific contributions.

\begin{enumerate}
 
 \item In Section~\ref{subsec:ucbm} we present an algorithm \UCBM, which
 solves \CRM, and achieves
 {$\R_T^* \in O(KM + {({K^{3/2}}/{M}})\sqrt{T\log({T}/{MK}}))$} over
 an unknown but finite horizon of $T$ pulls. The existing upper bound on regret
 is due to \cite{Liau+PSY:2018} and it involves problem specific quantities.
 Hence ours is the first problem independent finite time regret upper bound.
 Also, in Section~\ref{subsec:exptfinite} we empirically compare  \UCBM and its variations with the existing
 algorithms those solve \CRM.

  \item In Section~\ref{subsec:qrm2ucbmalg} we present a meta-algorithm \QRMUCBM,
 that uses \UCBM as a subroutine to the algorithm
 \QRMTWO~\citep{RoyChaudhuri+K:2018} to solve \QURM,
 and achieves $\R_T(\rho) \in o\left(\left(\frac{1}{\rho}\log\frac{1}{\rho}\right)^{4.9}\right. + M T^{0.205} + \left. T^{0.81}\sqrt{\frac{\log M}{M^2} \log \frac{T}{M}}\right)$. 
 In Section~\ref{subsec:exptinfinite} we experimentally demonstrate that \QRMUCBM (in terms of conventional regret $\R_T^*$) is more efficient than the algorithms by~\cite{Herschkorn+PR:1996}
 and~\cite{Berry+CZHS:1997}, on problem instances with Bernoulli arms.
\end{enumerate}
We briefly review the existing literature, before we present the key intuitions in Section~\ref{sec:kintuit}.

\section{Related Work}
\label{sec:relwork}

Started by \citet{Robbins:1952} the predominant body of literature in stochastic multi-armed bandit is 
dedicated to the regret minimisation task on finite and infinite bandit 
instances. Later, a number of salient algorithms like \UCB~\citep{Auer+CF:2002}, Thompson Sampling~\citep{Chapelle+L:2011,Agrawal+G:2012}, \Moss~\citep{Audibert+B:2009} has been shown to achieve
the order optimal cumulative regret on the finite instances. When the number of arms is infinite, 
algorithms make special assumption is made on the reward function~\citep{Agrawal:1995,Kleinberg:2005} or on 
the sampling distribution~\citep{Wang+AM:2008} to guarantee a sub-linear regret. Despite a thorough
study on the finite and the infinite instances, the number of investigations in the memory frugal algorithms is limited. 

  Finite memory hypothesis testing has been drawing the attention of researchers since 
  long~\citep{Robbins:1956,Isbell:1959,Cover:1976}. However, in MAB 
  setting \citet{Cover:1968} first presented a finite memory algorithm for two-armed Bernoulli instance, that 
  achieves an average reward which converges to the optimal proportion in the limit,  with probability 1. His 
  approach consisted of a collection of interleaved test and trial blocks, where each test block is divided into 
  several sub-blocks and the switching among these sub-blocks is governed by a finite state machine. However, he
  considered only two-armed Bernoulli instances, and the approach guarantees only an asymptotic convergence of the 
  empirical average reward. Hence, this setup is not very interesting, as our objective is to present a finite-time
  analysis of regret for general bandit instances.

  \citet{Herschkorn+PR:1996} presented the first \emph{non-recalling}  algorithm for infinite bandit instances with Bernoulli arms, that maximises the almost sure average reward over
  an infinite horizon. \citeauthor{Berry+CZHS:1997}~\yrcite{Berry+CZHS:1997} improved over them for the problem instances 
  where the sampling distribution $P_\A$ is uniform over the set expected rewards of the 
  Bernoulli arms. Towards relaxing the assumption of Bernoulli reward distribution 
  \citeauthor{Pekoz:2003}~\yrcite{Pekoz:2003} showed that a  peculiarity that may arise
    if the reward distributions of the arms are not stochastically-ordered.
    Specifically, for some function $f: \R^+ \mapsto \R^+$, he proposed two policies---PolicyA and PolicyB parameterised by $f(\cdot)$, where the latter is a non-stationary version of the former. Then he showed that for some choice of $f(\cdot)$,
 there exist instances with a bounded positive reward on which  in the limit,
 exactly for one of PolicyA and PolicyB, the average reward will converge to the supremum mean reward, while
 for the other, it will converge to the infimum mean reward.
  Most recently, \citeauthor{Liau+PSY:2018}~\yrcite{Liau+PSY:2018} have presented an \emph{explore-then-commit}
  strategy based algorithm \textsc{UCBConstantSpace} that incurs sublinear finite-time regret on any
  finite bandit instance. However, their algorithm explicitly uses PAC-based
  arm elimination strategy that leads to a high regret. On the other hand,
  like the previous algorithms, their algorithm does not have the provision to take the
  advantage availability of larger arm memory. 
  Next, we describe the key intuitions behind our approach.

\section{Key Intuitions} 
\label{sec:kintuit}
One of our objectives is to solve \CRM for finite
bandit instances ($|\A| < \infty$). The problem is interesting for $M < |\A|$;
otherwise, one can solve the problem by using any existing regret minimisation
algorithm like UCB1~\citep{Auer+CF:2002} \etc. Intuitively, any algorithm that
solves \CRM for finite bandit instances, must ensure that the probability of 
pulling the optimal arm is
increased by progressively increasing at least one of the 
two probabilities---first, the probability of the optimal arm $a^*$ is in
arm memory; second, if $a^*$ is in arm memory, the probability that it will be 
pulled more often than the other arms in arm memory. 
For any algorithm that achieves sub-linear regret we can write, for any horizon $T$,
$\R_T^* \in o(T) \implies \Lim_{T \uparrow \infty} \frac{\R_T^*}{T} = \Lim_{T \uparrow \infty} \frac{1}{T}\sum_{t=1}^T \Pr\{a_t = a^*\} = 1$. Now, imposing the arm memory constraint, and letting $X_t$ be the current arm memory at $t$-th pull, we notice,
$\{a_t = a^*\} \implies \{a^* \in X_t\}$. Therefore,
\begin{equation}
\label{eq:relregcrm}
\resizebox{0.9\columnwidth}{!}{
$\Pr\{a_t = a^*\}= \E_{X_t} \left[\Pr\{a_t = a^*| a^* \in X_t\} \Pr\{a^* \in X_t\}\right].$
}
\end{equation}

Hence, the \emph{necessary and the sufficient} condition for an algorithm that \emph{asymptotically} solves \CRM is 
\begin{equation*}
\resizebox{\columnwidth}{!}{%
 $\Lim_{T \uparrow \infty} \frac{\sum_{t=1}^T \E_{X_t} \left[\Pr\{a_t = a^*| a^* \in X_t\} \Pr\{a^* \in X_t\}\right]}{T} = 1,$
 }
\end{equation*}
for $|X_t| \leq M$, where $1 \leq t \leq T$.

Given a bandit instance, algorithm of \citeauthor{Liau+PSY:2018}~\yrcite{Liau+PSY:2018} first solves a pure exploration problem
for a horizon $\bar{T}$ (a function of the mean reward of the arms) to maximise the quantity 
$\Pr\{a^* \in X_t\}$ in the R.H.S. of Equation~\eqref{eq:relregcrm}.
Once the number of pulls crosses $\bar{T}$, it chooses the arm with the highest empirical reward
in $X_t$ as the contentious best arm, and assigns the rest of the  horizon to that arm. Therefore,
for $t > \bar{T}$, it switches to pure-exploitation mode, thus maximising 
the quantity $\Pr\{a_t = a^* | a^* \in X_{\bar{T}}\}$. On the contrary, 
we adopt balanced exploration with an
aim to simultaneously increase $\Pr\{a^* \in X_t\}$ and $\Pr\{a_t = a^* | a^* \in X_{t}\}$.
Therefore, our algorithm does not depend on such $\bar{T}$.
It should be noted that, for the sake of sufficient exploration, an algorithm
should not stick to the same arm memory for too long, however, while selecting 
new arms (not in the current arm memory), it must judiciously choose the 
in memory arms to
replace. This trade-off relies on the notion of
\emph{simple regret} which we introduce next.

\paragraph{Simple Regret.} 
 Whereas the cumulative regret minimisation problem is based on the 
trade-off between exploration and exploitation, there is a separate line of 
literature in \emph{pure exploration} setting. One of the popular problems 
in pure exploration setting is to minimise ``simple regret''. If $b_t \in \A$ is
 the arm recommended by the
 algorithm after the $t$-th pull, then the
 simple regret of the algorithm at $t$ is defined as,
  \begin{equation}
   \label{eq:simpleregdef}
   \E[r^*] \defeq \mu^* - \E[\mu_{b_t}].
  \end{equation}

\paragraph{Relation of Cumulative Regret Minimisation with the Minimisation of Simple Regret.} 
 \citet{Bubeck+MS:2009} gave a general definition of a forecaster, depicted in 
 Figure~\ref{fig:genforecaster}.
 Given a set of arms as $\A$ as input,
 at each step $t$, possibly depending on $H_{t-1}$, it
 selects an arm $a_t$ by using a strategy called \emph{``allocation strategy''}.
 On pulling the arm $a_t$ it receives a reward $r_t$,
 and executes a \emph{``recommendation strategy''} that takes $H_t$ as the input and
 outputs an arm $b_t$. The forecaster continues to alternately execute 
 \emph{allocation strategy} and \emph{recommendation strategy} until some stopping
 condition is met.
 
\begin{figure}[h]
    \centering
    \resizebox{\columnwidth}{!}{\footnotesize
     \adjustbox{minipage=[r]{0.9\linewidth},scale={1},margin=0.5em,frame,center} {
	$t = 1$, $H_0 = \{\emptyset\}$.\\
	\textbf{While} (stopping condition is not met)\{
	  \begin{enumerate}
	   \item Execute \emph{allocation strategy} that possibly depending on 
	      $H_{t-1}$ selects an arm $a_t$.
	   \item Pull the arm $a_t$, and get a reward $r_t$. Update $H_t = H_{t-1} \cup \{(a_t, r_t)\}$.
	   \item Execute \emph{recommendation strategy} that  possibly depending on 
	      $H_{t}$ outputs an arm recommendation $b_t$. Update $t = t+1$.
	    \}
	  \end{enumerate}
      }
      }
    \caption{A general forecaster.}
    \label{fig:genforecaster}
\end{figure}

  A careful look reveals that 
  a forecaster that
  at each step, $t$, recommends the arm, which is selected by allocation 
  strategy in the same step
  (that is $b_t \equiv a_t$ in Figure~\ref{fig:genforecaster}), then the
  cumulative regret (Equation~\ref{eq:cumregdef}) of that forecaster
  is identical to the sum of simple regret (Equation~\ref{eq:simpleregdef})
  over time steps $t$. This tempts one to intuit that using an
  allocation strategy which incurs low cumulative
  regret will help in designing a forecaster that achieves a small
  simple regret and vice-versa. However,
  \citeauthor{Bubeck+MS:2009}~\yrcite{Bubeck+MS:2009} present a negative  result on this trade-off. 
  Further they \citep{Bubeck+MS:2009}
  present a upper bound on $\E[r^*]$ for a number of 
  forecasters~\citep[Table 1]{Bubeck+MS:2009} one of which is defined below:
    
  \begin{definition}[\UCBMPA.]
    A forecaster, which at each step uses UCB1~\citep{Auer+CF:2002}
    as the
    allocation strategy, and uses the recommendation strategy that
    outputs the \emph{\underline{m}ost \underline{p}layed \underline{a}rm}
    (MPA), shall be called as \UCBMPA.
  \end{definition}

  We quote their result \citep[Theorem 3]{Bubeck+MS:2009} in Theorem~\ref{thm:bubeckucbmpa}
  which serves as the cornerstone of analysis of the algorithm \UCBM.
  
  \begin{theorem}[Distribution-free upper bound on Simple-Regret of \textsc{UCB-MPA} by~\citet{Bubeck+MS:2009}]
   \label{thm:bubeckucbmpa}
   Given a $K$-sized set of arms $\A$ as input, if \UCBMPA runs for a
   horizon of $T$ pulls
   such that $T \geq K(K+2)$, then for some constant $C > 0$, 
   it achieves the expected simple regret $\E[r^*] \leq C \sqrt{\frac{K\log T}{T}}$.
  \end{theorem}

Although \UCB was originally designed as a cumulative regret minimisation algorithm, empirically it performs
reasonably well as an exploration strategy to give us a good balance between exploration and exploitation.
We choose \UCBMPA over other forecasters as it is easy to comprehend and leads a simpler
derivation. 
For the rest of the paper
we shall use $\log$ and $\ln$ to denote base 2 and natural logarithm respectively.
Also, for any positive integer $Z$ we shall denote the set $\{1, 2, 3, \cdots, Z\}$ by $[Z]$.


\section{Algorithm for Finite Bandit Instances}
\label{sec:algfinite}

We present the algorithm  \UCBM and establish a problem-independent upper-bound on the cumulative regret. 
Then we empirically compare \UCBM and its variations with the algorithm by \citet{Liau+PSY:2018}.
Algorithm~\ref{alg:ucbmf} is based on \UCBMPA. However, 
one can the replace the underlying call to \UCB with any other allocation strategy 
like Thompson sampling~\citep{Agrawal+G:2012}, or
\textsc{Moss}~\citeauthor{Audibert+B:2009}~\yrcite{Audibert+B:2009}, as we do in our experiments.

  \subsection{Algorithm and Regret-Analysis.}
  \label{subsec:ucbm}
  Algorithm~\ref{alg:ucbmf} describes \UCBM that solves the problem \CRM for finite
  bandit instances. We
  improve upon the contribution of \citeauthor{Liau+PSY:2018}~\yrcite{Liau+PSY:2018} in three aspects---first,
  \UCBM  is empirically much more efficient even if we allow $M= 2$ (as opposed to $M = 4$ for theirs) as 
  it does not explicitly use
  pure exploration based elimination; second, it scales with the arm memory size;
  third, we present a distribution-free upper bound on the incurred regret of \UCBM for
  solving \CRM on finite bandit instances. 

  Given a finite set of arms $\A$ ($|\A| = K < \infty$), and arm memory size $M$ ($2 \leq M < K$)
  \UCBM approaches in phases.
  It breaks each phase into $h_0 \defeq \ceil{(K-1)/(M-1)}$ sub-phases.
  Inside any phase $w$, at each sub-phase $j$, it runs \UCBMPA on an $M$-sized subset
  of arms $S^{w,j}$ (called arm memory), and assigns the recommended arm to 
  $\hat{a}$, and forwards it to the next sub-phase.
  On the subsequent sub-phase (that might belong to the next phase),
  it chooses $M-1$ new arms from $\A$, along with the arm $\hat{a}$
  forwarded from the previous sub-phase,
  and repeat the previous steps. It is to be noted that the horizon spent on each sub-phase of a phase $w$ is the same and is given by $b_w$. Also,
  for $w \geq 2$, the total horizon spent in phase $w$ is given by $h_0 b_w = 2 h_0 b_{w-1}$.
  To satisfy the assumption in Theorem-3 of \citeauthor{Bubeck+MS:2009}~\yrcite{Bubeck+MS:2009},
  at the first phase, for each of the sub-phases, \UCBM chooses a horizon of
  $b_1 = M(M+2)$ pulls.
  For the rest of the analysis of \UCBM, we shall denote, the number of phases executed by \UCBM as $x_0$.

  For $M >= K$, as the arm memory size is large enough, 
  it is effectively removing the memory constraint. 
  In such \emph{unconstrained} scenarios it is preferable run
  \UCB~\citep{Auer+CF:2002} on the whole instance, which will incur a lower regret. We adopt this into \UCBM, and hence, running \UCBM with $M \geq K$ is identical to run original \UCB.
  

  
  \begin{algorithm}[h]
  \begin{algorithmic}
  {\footnotesize
  \REQUIRE {$\A$: the set of $K$ arms indexed by $[K]$, $M (\geq 2)$: Arm memory size.}
  \IF{$M \geq K$}{
    \STATE Run \protect\UCB on $\A$ until the horizon runs out.
    }
  \ELSE{
    \STATE $b_1 = M (M+2)$.\COMMENT{Initial horizon per sub-phase}
    \STATE $\hat{a} = 1$. \COMMENT{Initial arm recommendation}
    \STATE $w = 1$. \COMMENT{Counts number of phases}
    \STATE $h_0 = \bceil{\frac{K-1}{M-1}}$.\COMMENT{The number of sub-phases in a phase}
    \STATE Randomly shuffle the arm indices.
    \WHILE{the horizon is not finished}{
      \STATE $l = 0$
      \FOR{$j =1, \cdots, h_0$; if the horizon is not finished}{
	\STATE $S^{w,j} = \{l+1, \cdots, \min\{l+1 + (M-1), K\}\}$.
	\STATE $l \defeq$ The highest arm index in $S$.
	\IF{$\hat{a} \not\in S^{w,j}$}{
	  \STATE $S^{w,j} = \{\hat{a}\} \cup S^{w,j}\setminus\{l\}$.
	  \STATE $l = l -1$.
	  }\ENDIF
	  
	\COMMENT{\textsc{Allocation Strategy}}
	\STATE Run \protect\UCB on $S^{w,j}$ for horizon of $b_w$ pulls or the remaining horizon; whichever is smaller.
	\COMMENT{\textsc{Recommendation Strategy}}
	\STATE $\hat{a} \defeq$ The \emph{most played arm} in $S^{w,j}$.
	}\ENDFOR
      \STATE $w = w+1$. \COMMENT{Increment phase count}
      \STATE $b_w = 2 \cdot b_{w-1}$. \COMMENT{Increment horizon per sub-phase}
      }\ENDWHILE
  }\ENDIF
  }
  \end{algorithmic}
  \caption{\protect\UCBM: For finite bandit instances.}
  \label{alg:ucbmf}
  \end{algorithm}
  
 \begin{theorem}
  \label{thm:ubucbmpaf}
   Given a set of $K$ arms $\A$, with $K \geq 3$, an arm memory size $M$ such that $2 \leq M < K$, as input,
   then for a horizon of $T$ pulls, with $T > KM^2 (M+2)$,
   \UCBM will incur 
   $\R_T^* = O\left(KM + ({K^{3/2}}/{M})\sqrt{T\log({T}/{KM}})\right).$
  \end{theorem}
  
  We note that for a given bandit instance with $K$ arms, an arm memory size $M$,
  and horizon $T$, 
  the number of sub-phases is $h_0 = \ceil{(K-1)/(M-1)}$,
  and the total number of phases is upper bounded 
  by $x_0 \defeq \ceil{\log({2T}/{MK})}$ (Lemma~\ref{lem:maxnumphase} in Appendix~\ref{app:algfinite}).
  Now, we upper bound
  the maximum regret incurred in a sub-phase inside a given phase,
  and sum over all the sub-phases.

 As \UCBM ensures inclusion of the optimal arm at least once in every phase, we
 note the following.
  \begin{corollary}
   \label{cor:maxnsubphase}
    Let us denote, the sequence of sub-phase-wise arm memory as
   $\mathcal{S} \defeq \{S^{1,1}, S^{1,2}, \cdots,$ $S^{1,h_0}, S^{2,1}, S^{2,2},$
   $\cdots, S^{2,h_0},\cdots,S^{x_0,1},$ $S^{x_0,2}, \cdots, S^{x_0,h_0}\}$. Then,
   for $d \geq h_0$, at least one of any $d$ consecutive elements of $\mathcal{S}$
   contains $a^*$.
  \end{corollary}
   In any given phase, we need to upper-bound the difference of mean of the best in memory arm between two
   successive sub-phase. 
   Considering $\mathcal{S}$ as defined in Corollary~\ref{cor:maxnsubphase}, let $a_*^{y,j} \in S^{y, j} \in \mathcal{S}$  be the arm recommended by the sub-phase $j-1$ to $j$.
   It is important to note that
   $\max_{a \in S^{y,j}}\mu_a \geq \mu_{a_*^{y,j}}$.
   Therefore, in the interest of finding a upper bound on the regret, 
   it is safe to consider $\mu_{a_*^{y,j}} = \max_{a \in S^{y,j}}\mu_a $ as a pessimistic
   estimate of the best mean in $S^{y,j}$.
   In any given sub-phase $j \leq h_0 -1$ in phase $y$, we let $\E[r^{y}] \defeq \E[\mu_{a_*^{y,j}} - \mu_{a_*^{y,j+1}}]$. Now, noticing that  on each sub-phase in a phase $y$
   \UCBM spends $b_y$ pulls we upper bound $\E[r^{y}]$ as follows.
   \begin{corollary}
   \label{cor:succsimpreg}
   Using Theorem~\ref{thm:bubeckucbmpa}, in phase $y$, at the end 
   of each sub-phase $j$, the expected simple regret with respect to $\mu_{a_*^{y,j}}$ is upper-bounded as 
   $\E[r^{y}] \leq C \sqrt{(M \log b_y)/b_y}$. The upper bound is independent of $j$ as the budget for each sub-phase in a given phase remains the same.
   \end{corollary}

    We notice, that the arm forwarded from each sub-phase to the next one, not necessarily to be the optimal arm. Hence, in the worst case, the expected difference between 
  the mean of the optimal arm, and the highest mean reward in the current arm memory grows 
  linearly with the number of sub-phase in a given phase. We upper bound it as follows.
    \begin{restatable}{lemma}{lemubsimpreg}
   \label{lem:ubsimpreg}
   Let, we are given a $K$-sized set of arms $\A$, and an arm memory size $M$.
   Also, let as defined in Corollary~\ref{cor:succsimpreg} at any phase $y\geq 2$, in the sub-phase $j$,
   if $\mu_*^{y,j}$ is the maximum of the mean of the arms in the arm-memory,
   then $$\max_{1\leq j \leq h_0} \E[\mu^* - \mu_*^{y,j}] \leq 2h_0 \E[r^{y}].$$
  \end{restatable}
    The proof is presented in Appendix~\ref{app:algfinite}. Next, we use Lemma~\ref{lem:ubsimpreg}
    to upper bound the cumulative regret ($\R_T^*$).

 
 \paragraph{Bifurcation of $\R_T^*$.} 
  For any given phase $w$, and a sub-phase $j$, let $\mu_*^{w,j} \defeq \max{\mu_a: a\in S^{w,j}}$,
  and $R_{w,j}$ be the incurred regret. Then, 
 {\footnotesize
  \begin{align*}
   & R_{w,j}^* = b_w \mu^* - \sum_{i=1}^{b_w}\E[\mu_{a_t}]\\
   & = b_w \E[\mu^* - \mu_*^{w,j}] + \sum_{t=1}^{b_w}(\E[\mu_*^{w,j}]  - \E[\mu_{a_t}]).
  \end{align*}
 }
  Where the expectation is taken over all possible sources of randomisation.
  Now, letting $R_{w,j}^{(1)} = b_w (\mu^* - \mu_*^{w,j})$, and $R_{w,j}^{(2)} = \sum_{t=1}^{b_w}(\E[\mu_*^{w,j}]  - \E[\mu_{a_t}])$,
  we can write,
  \begin{equation}
  \label{eq:breakcumreg}
   R_T^* = \sum_{w=1}^{x_0} \sum_{j=1}^{h_0} R_{w,j}^* = \sum_{w=1}^{x_0} \sum_{j=1}^{h_0} (R_{w,j}^{(1)} + R_{w,j}^{(2)}).
  \end{equation}
    
    Now, using Lemma~\ref{lem:ubsimpreg}  we  upper bound $R_{w,j}^{(1)}$ as follows.
    
%
  \begin{restatable}[]{lemma}{lemsumrswj}
   \label{lem:sumrswj}
   For $2 \leq M < K$, and  for $T > KM^2(M+2)$, and for some constant $C'$, 
   $\sum_{w=1}^{x_0}\sum_{j=1}^{h_0} R_{w,j}^{(1)}$ $\leq C'$ $\left(KM + ({K^{3/2}}/{M})\sqrt{T\log({T}/{MK}})\right)$.
  \end{restatable}
  For the detailed proof we refer to Appendix~\ref{app:algfinite}. We note, that 
   $\sum_{w=1}^{x_0}\sum_{j=1}^{h_0} R_{w,j}^{(2)}$, can be upper-bounded using 
   the problem independent upper-bound on the cumulative regret of \UCB~\citep{Auer+CF:2002}, as we restate below.

  \begin{lemma}[Distribution-Free Upper Bound on Cumulative Regret of UCB1~\citeauthor{Auer+CF:2002}~\yrcite{Auer+CF:2002}]
  \label{lem:ucb1ub}
   Given a set of $K$-arms as the input, for any horizon $T$, the cumulative regret
   incurred by UCB1 $R_T^* \leq 12\sqrt{TK\log T} + 6K$. Further, if $T \geq K/2$,
   then $R_T^* \leq 18\sqrt{TK\log T}$.
  \end{lemma}

 Next, using Lemma~\ref{lem:ucb1ub}, we upper bound $\sum_{w=1}^{x_0}\sum_{j=1}^{h_0} R_{w,j}^{(2)}$, 
 with the proof given in Appendix~\ref{app:algfinite}.
  \begin{restatable}[]{lemma}{lemsumrcwj}
   \label{lem:sumrcwj}
   For $2 \leq M < K$, and  $T > KM^2 (M+2)$, and for some constant $C''> 0$, 
   $\sum_{w=1}^{x_0}\sum_{j=1}^{h_0} R_{w,j}^{(2)}\leq$ $C'' \left(KM + \sqrt{TK \log ({T}/{MK}})\right)$.
  \end{restatable}

  \paragraph{Proof of Theorem~\ref{thm:ubucbmpaf}}
   Using Equation~\ref{eq:breakcumreg}, and applying Lemma~\ref{lem:sumrswj} and Lemma~\ref{lem:sumrcwj}
   we prove the theorem.
  
Next, we present an empirical comparison of \UCBM and some of its variations with the algorithm of \citet{Liau+PSY:2018}.

\subsection{Experiment}
\label{subsec:exptfinite}
The use of \UCB~\citep{Auer+CF:2002}
as a subroutine in Algorithm~\ref{alg:ucbmf}, can be replaced by 
any other allocation strategies, which in effect will give rise to a different
upper bound. In the interest of studying the empirical behaviour,
we  consider \Moss \citep{Audibert+B:2009} and \TS \citep{Agrawal+G:2013}
in our experiments, and rename \UCBM to \TSM and \MSM respectively.
However, everything else (in Algorithm~\ref{alg:ucbmf})
including the recommendation strategy, is kept unchanged.

\paragraph{Bandit Instances.} We run the experiments on three different instances.
Let, $K$ be the number of arms in the instance. Also, for convenience, let
the arms indices be sorted in descending order of their mean, with
$\mu_1 = \mu^* = 0.99$. As we randomly permute the arm indices in
all our experiments, this assumption does not affect the results.
We write $\B^K_{L}$ to denote an instance in which the mean of the $K$ arms are
linearly spaced between $0.99$ $(=\mu^*)$ and $0.01$. 
The other two $K$-armed instances which are analogous to the ones used by~\citeauthor{Jamieson+MNB:2014}~\yrcite{Jamieson+MNB:2014}.
For $\alpha \in \{0.3, 0.6\}$, they are defined as $\B^K_{\alpha}$,
in which any sub-optimal arm $i > 1$, has the mean
$\mu_i = 0.01 + \mu^* - (\mu^*-0.01)((i-1)/ (K-1))^\alpha$.

\begin{figure}[ht]
 \centering
 \includegraphics[height=1in,width=0.8\columnwidth]{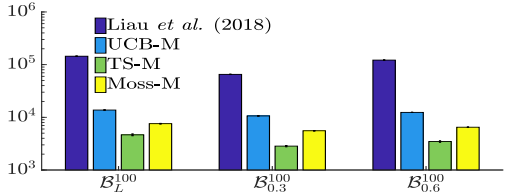}
 \caption{Comparison of incurred cumulative regret 
	     in $\log$ scale (y axis), by
	     \protect\UCBM, \protect\TSM, \protect\MSM, for $M=2$, and
	     the algorithm of \citeauthor{Liau+PSY:2018}~\yrcite{Liau+PSY:2018}
	     (uses $M = 4$) after
	     $10^6$ pulls. Each bar represents the average over 100 iterations, and
	     with one standard error on mean. For details about the instances and the
	     algorithms we refer to Section~\ref{subsec:exptfinite}.}
  \label{fig:k100reg}
\end{figure}

For $K=100$, Figure~\ref{fig:k100reg}
compares the cumulative regret incurred by 
algorithms \UCBM, \TSM, \MSM for an arm memory size $M=2$,
with the algorithm of \citeauthor{Liau+PSY:2018}~\yrcite{Liau+PSY:2018} (\textsc{UCBConstantSpace}). 
A comparison of cumulative regret, and the number of pulls to the individual arms
in the instances with $K=10$ is presented in Figure~\ref{fig:app:k10} in 
Appendix~\ref{app:experiment_finite}. It is important to note that despite using larger arm-memory of $M =4$,
which is twice of the others, their algorithm
incurs a significantly higher regret.

Intuitively, \citeauthor{Liau+PSY:2018}'s~\yrcite{Liau+PSY:2018} algorithm first solves a pure exploration 
problem to identify a near optimal arm, and then commits
the rest of the horizon to that arm.
Consequently, it spends a prohibitively large number of pulls on the sub-optimal
arms leading to a high regret. In contrast, we just make sure that at any instant,
the expected difference between the mean of the
optimal arm and the best arm in the current arm memory is not too large. 
Apparently, this difference increases with the subsequent sub-phases. However,
\UCBM ensures to choose the optimal arm in its arm memory at least once
in any given phase leading to a ``reset" to this difference. 
On the other hand, this difference progressively reduces due to
doubling the budget in each phase.
This explains why \UCBM, \TSM, and \MSM incur significantly lower regret.

As \UCBM can take advantage of larger arm memory size, next we shall compare the
incurred regret by varying it.
Recalling the algorithm \UCB~\citep{Auer+CF:2002}, if an arm $a$ has been pulled $u_a^t$ times
till the time step $t$, and if $\hat{\mu}_a^t$ is its empirical average reward, then the upper
confidence bound of that arm is given by 
$ucb_a^t = \hat{\mu}_a^t + \eta \sqrt{2\log t/u_a^t},$ with $\eta = 1$. 
It can be experimentally validated that tuning $\eta$ can lead to achieving a smaller
regret as claimed by the authors~\citep[Section 4]{Auer+CF:2002}. 
We present the regret incurred by \UCBM for $\eta = 0.2$, alongside the other
algorithms.

\begin{figure}[ht]
    \centering
    \includegraphics[height=1in,width=0.8\columnwidth]{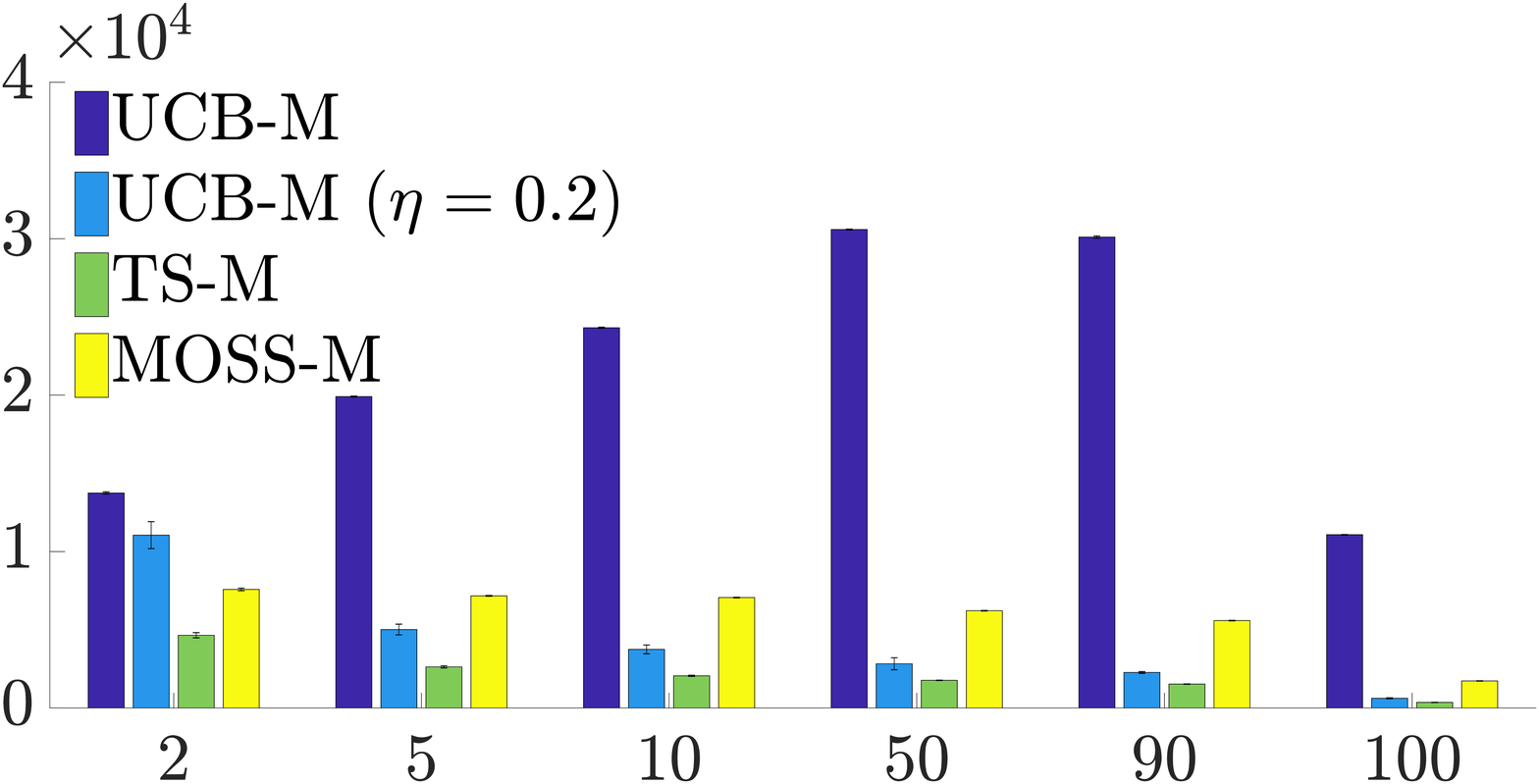}
     \caption{Comparison of incurred regret (y axis) on the instance $\B_L^{100}$
 by different algorithms by varying arm memory size $M$ (x axis), and the horizon.
 Each bar represents incurred regret averaged over 100 iterations, and
	     with one standard error. For details about the instances and the
	     algorithms we refer to Section~\ref{subsec:exptfinite}.}
    \label{fig:varmem_B100L_h106}
\end{figure}

Intuition suggests that increasing arm memory should help in achieving a low
regret, as it increases the chance of pulling the optimal arm more frequently.
Also, the upper bound given by Theorem~\ref{thm:ubucbmpaf} supports this intuition.
However, in practice, we notice an interesting behaviour.
On the instance $B_{L}^{100}$, we compare the cumulative regret incurred
by \UCBM, \TSM, and \MSM in Figure~\ref{fig:varmem_B100L_h106} by varying $M$. 
For a comparison on the
other instances, the reader is referred to Figure~\ref{fig:varmem_B100A03}
in Appendix~\ref{app:experiment_finite}.
As expected, \UCBM, \TSM and \MSM always incur a higher regret than their 
unconstrained ($M = K$) counter parts. Also, for \UCBM with $\eta = 0.2$,
\TSM and \MSM increasing the arm 
memory size $M$ makes them achieve a lower regret. However, the behaviour of 
\UCBM ($\eta = 1$) is significantly different from the other two. If $M < K$, it incurs a 
relatively low 
regret for $M=2$. Afterwards it increases with $M$, followed by a slow decrease.
We conclude that this peculiarity in its behaviour is due to the intrinsic looseness in the calculation of upper confidence bound. Also, that is the reason
why \UCBM with $\eta = 0.2$, and the others not only incur a lower regret but behave consistently.

\section{Algorithm for Infinite Bandit Instances}
\label{sec:qrm2ucbm}
In this section, we provide a bounded arm-memory algorithm and its upper bound on the incurred quantile-regret. Also, on various problem instances 
we empirically compare 
its incurred conventional cumulative regret ($\R_T^*$) with the existing algorithms.

\subsection{Algorithm and Quantile-Regret Analysis}
\label{subsec:qrm2ucbmalg}
We solve the problem of \QURM by modifying the algorithm
\QRMTWO~\citep{RoyChaudhuri+K:2018} to make it
use \UCBM as the sub-routine, and adjust the arm exploration rate
accordingly to minimise the upper bound. We call it \QRMUCBM and describe in Algorithm~\ref{alg:qrm2ucbm}.

 \begin{algorithm}[h]
  \begin{algorithmic}
  {\footnotesize
  \REQUIRE{$\A, P_\A$, and arm memory size $M$}
   \STATE Set $\alpha = 0.205$, $B = \left(M^2 (M+2)\right)^\frac{1}{1-\alpha}$, and $\K_0 = \emptyset$
   \FOR{$r = 1, 2, 3, \cdots$}{
     \STATE $t_r = 2^r B, n_r = \lceil t_r^\alpha \rceil$.
     \STATE Form a set $\K_r$ by selecting additional $n_r - |\K_{r-1}|$ arms from $\A$ using $P_\mathcal{A}$, and adding to $\K_{r-1}$.
     \STATE Run \UCBM on $\K_r$, for horizon of $t_r$, with arm memory size $M$
     }\ENDFOR
  }
  \end{algorithmic}
  \caption{\protect\QRMUCBM}
  \label{alg:qrm2ucbm}
 \end{algorithm}

 Below, we present the upper bound on the quantile regret incurred 
 by \QRMUCBM.
 \begin{theorem}[Sub-linear quantile-regret of \protect\QRMUCBM]
 \label{thm:ubqrm2ucbm}
 For $\rho \in (0, 1)$ and for sufficiently large $T$, \QRMUCBM incurs 
\begin{equation*}
    \resizebox{\columnwidth}{!}{%
    $R_T(\rho) \in o\left(\left(\frac{1}{\rho}\log\frac{1}{\rho}\right)^{4.89} + M T^{0.205} + T^{0.81}\sqrt{\frac{\log M}{M^2} \log \frac{T}{M}}\right)$.
    }
\end{equation*}
 \end{theorem}
 \begin{proof}
 
To prove the theorem we follow the steps of proof for Theorem 3.3 in \citeauthor{RoyChaudhuri+K:2018}~\yrcite{RoyChaudhuri+K:2018}.
For  any fixed  $\rho \in (0, 1)$, we break the analysis for upper bound
on $\R_T(\rho)$ in cases---first, the algorithm never encounters an arm from
$\mathcal{TOP}_\rho$; second, it picks at least one arm from $\mathcal{TOP}_\rho$.
 
 The key step in the analysis of the first part is showing that there exists $r^{*} \geq 1$ such that for all $r \geq r^{*}$, the set of arms $\K_r$ is sufficiently large to contain an arm from $\mathcal{TOP}_\rho$ with high probability. Defining $\mathcal{TOP}_\rho$ is in $\K_r$ as $E_r(\rho) \defeq \{\K_r \cap \mathcal{TOP}_\rho = \emptyset\}$, and following the
 steps for the derivation of Equation (3) in the proof of Theorem 3.3 in \citeauthor{RoyChaudhuri+K:2018}~\yrcite{RoyChaudhuri+K:2018} we 
 arrive at
 
\begin{equation}
\label{eq:regempty2}
\resizebox{0.9\columnwidth}{!}{%
$\sum_{r=1}^{\log T} t_r \Pr\{E_r(\rho)\} \in O \left(\left(\frac{1}{\rho}\log\frac{1}{\rho}\right)^\frac{1}{\alpha} +  T^{1-\frac{\alpha\log\mathrm{e}}{1+\gamma}}\right)$.
}
\end{equation}
The detailed derivation of Equation~\eqref{eq:regempty2} is given in 
Lemma~\ref{lem:qreger} in Appendix~\ref{app:qrm2ucbm}.

In the second part, we upper-bound the incurred regret  for the case where 
\QRMUCBM encounters at least one arm from the $\mathcal{TOP}_\rho$ in $\K_r$
(the event $\neg E_r(\rho))$. Using Theorem~\ref{thm:ubucbmpaf} and using the similar approach
for deriving Equation~(4) in the proof of Theorem 3.3 in \citet{RoyChaudhuri+K:2018} we 
arrive at
{\footnotesize
\begin{align}
\label{eq:regnonempty}
& \sum_{r=r^{*}}^{\log T} C \left(n_r M + ({n_r^{3/2}}/{M})\sqrt{t_r \log t_r}\right),\nonumber\\
  & \leq C'\left(M T^{\alpha} + ({\sqrt{\log M}}/{M})\sqrt{T^{1+3\alpha}\log({T}/{M}})\right),
\end{align}}
for some constant $C'$. The intermediate steps to obtain
\eqref{eq:regnonempty} are presented in Lemma~\ref{lem:qregne} in Appendix~\ref{app:qrm2ucbm}. Combining Equation~\eqref{eq:regempty2}, Equation~\eqref{eq:regnonempty}
and substituting for $t_{r^*}$,
the upper bound on $\R_T(\rho)$ \wrt $T$ gets minimised for $\alpha = 1/(3 + 2\log\mathrm{e}/(1+\gamma)) \approx 0.205$, thus proving the theorem.
 \end{proof}
It is to be noted that
inside \QRMUCBM one can use the algorithm by \citet{Liau+PSY:2018} instead of \UCBM. However,
as we have already shown in Section~\ref{subsec:exptfinite} that \UCBM is empirically superior than
their algorithm, we do not consider this variation in our our experiment.

\subsection{Experiment}
\label{subsec:exptinfinite}

Although \QRMUCBM is designed with the aim to minimise quantile-regret, we
use conventional cumulative-regret as the evaluation metric.
Similar to \UCBM, the algorithm \QRMUCBM can be altered to use 
\TSM or \MSM as the subroutine instead, and we call them \QRMTSM and  \QRMMSM 
respectively. Algorithm~\ref{alg:qrm2ucbm} uses the value of $\alpha$ that minimises the 
upper bound on regret in Theorem~\ref{thm:ubqrm2ucbm}. However, for empirical efficiency, we keep the
$\alpha=0.347$ as used by the algorithm \QRMTWO~\citep{RoyChaudhuri+K:2018}.
We compare incurred conventional regret by each of these algorithms
against the algorithms by \citeauthor{Herschkorn+PR:1996}~\yrcite{Herschkorn+PR:1996} and \citeauthor{Berry+CZHS:1997}~\yrcite{Berry+CZHS:1997},
and present it in Table~\ref{tab:compare_herschberry_a0347}.
We use the same four Bernoulli instances used by \citet{RoyChaudhuri+K:2018}---instances $I_1$ and $I_2$ have $\mu^*=1$,
and the probability distributions on $\mu$ induced by $P_\A$ are given by $\beta(0.5,2)$, and $\beta(1,1)$ respectively. Similarly,
instances $I_3$ and $I_4$ have $\mu^*=0.6$,
and the  probability distributions on $\mu$ induced by $P_\A$ are given by scaled $\beta(0.5,2)$, and $\beta(1,1)$ respectively.
Each column of the tables
is labelled by the corresponding probability density function of encountering the mean rewards.
As Table~\ref{tab:compare_herschberry} suggests, the existing 
algorithms incur a significantly higher regret in most of the cases.
We put the comparison for $\alpha=0.205$ at Table~\ref{tab:compare_herschberry} in Appendix~\ref{app:experiment_infinite}.

It is interesting to note that, like the finite instances, increasing arm memory leads to a lower regret.
Specifically, the scaled version of \QRMUCBM  (using \UCBM with $\eta=0.2$) along with \QRMTSM and \QRMMSM show an improvement with larger arm memory. However, with $\eta=1$ in the underlying \UCBM, \QRMUCBM fails to take
the advantage of larger arm memory.

\begin{table}[]
\centering
\caption{Cumulative regret ($/10^{5}$) of \protect\QRMUCBM, \protect\QRMTSM, \protect\QRMMSM (with $\alpha=0.347$) and the strategies proposed by \citeauthor{Herschkorn+PR:1996}~\yrcite{Herschkorn+PR:1996} and \citeauthor{Berry+CZHS:1997}~\yrcite{Berry+CZHS:1997} after $10^6$ pulls, on instances $I_1$, $I_2$, $I_3$ and $I_4$. Each result is the average of 20 runs, showing one standard error.}
\label{tab:compare_herschberry_a0347}
\resizebox{\columnwidth}{!}{%
\begin{tabular}{|l|c|l|l|l|l|}
\hline
\multicolumn{1}{|c|}{Algorithms} & M & \multicolumn{1}{c|}{\begin{tabular}[c]{@{}c@{}}$I_1$: $\beta(0.5,2)$\\ $\mu^*=1$\end{tabular}} & \multicolumn{1}{c|}{\begin{tabular}[c]{@{}c@{}}$I_2$: $\beta(1,1)$\\ $\mu^*=1$\end{tabular}} & \multicolumn{1}{c|}{\begin{tabular}[c]{@{}c@{}}$I_3$: $\beta(0.5,2)$\\ $\mu^*=0.6$\end{tabular}} & \multicolumn{1}{c|}{\begin{tabular}[c]{@{}c@{}}$I_4$: $\beta(1,1)$\\ $\mu^*=0.6$\end{tabular}} \\ \hline
\begin{tabular}[c]{@{}l@{}}Non-stationary Policy\\ \citep{Herschkorn+PR:1996}\end{tabular} & 1 & 3.58 $\pm$0.4 & 1.11 $\pm$0.2 & 1.64 $\pm$ 0.2 & 0.79 $\pm$ 0.1 \\ \hline
\begin{tabular}[c]{@{}l@{}}$\sqrt{T}$-run\\ \citep{Berry+CZHS:1997}\end{tabular} & 2 & 6.18$\pm$0.5 & 1.11$\pm$0.4 & 4.18$\pm$0.3 & 2.03$\pm$0.3 \\ \hline
\begin{tabular}[c]{@{}l@{}}$\sqrt{T}\ln T$-learning\\ \citep{Berry+CZHS:1997}\end{tabular} & 2 & 6.32$\pm$0.4 & 0.69$\pm$0.3 & 4.38$\pm$0.2 & 2.15$\pm$0.3 \\ \hline
\begin{tabular}[c]{@{}l@{}}Non-recalling $\sqrt{T}$-run\\ \citep{Berry+CZHS:1997}\end{tabular} & 1 & 5.35 $\pm$0.5 & \textbf{0.03} $\pm$0.004 & 4.56 $\pm$ 0.001 & 2.55 $\pm$ 0.001 \\ \hline
\rule{0pt}{4ex}
\multirow{2}{*}{\textsc{QUCB-M}} & $2$ & 1.84$\pm$0.17 & 0.41$\pm$0.02 & 1.29$\pm$0.10 & 0.49$\pm$0.02 \\ \cline{2-6} 
\rule{0pt}{4ex}
 & $10$ & 1.98$\pm$0.16 & 0.59$\pm$0.02 & 1.49$\pm$0.09 & 0.63$\pm$0.01 \\ \hline
\rule{0pt}{4ex}
\multirow{2}{*}{\textsc{QUCB-M} ($\eta = 0.2$)} & $2$ & 2.00$\pm$0.20 & 0.32$\pm$0.05 & 1.41$\pm$0.10 & 0.69$\pm$0.04 \\ \cline{2-6} 
\rule{0pt}{4ex}
 & $10$ & 1.71$\pm$0.16 & 0.16$\pm$0.02 & 1.16$\pm$0.09 & {0.30}$\pm$0.02 \\ \hline
\rule{0pt}{4ex}
\multirow{2}{*}{\textsc{QTS-M}} & $2$ & 1.77$\pm$0.17 & 0.32$\pm$0.04 & 1.23$\pm$0.09 & 0.40$\pm$0.02 \\ \cline{2-6} 
\rule{0pt}{4ex}
 & $10$ & 1.91$\pm$0.16 & 0.18$\pm$0.03 & 1.14$\pm$0.10 & {0.30}$\pm$0.02 \\ \hline
 \rule{0pt}{4ex}
\multirow{2}{*}{\textsc{QMoss-M}} & $2$ & 1.74$\pm$0.17 & 0.31$\pm$0.02 & 1.20$\pm$0.10 & 0.39$\pm$0.02 \\ \cline{2-6} 
\rule{0pt}{4ex}
 & $10$ & \textbf{1.69}$\pm$0.15 & 0.25$\pm$0.02 & \textbf{1.13}$\pm$0.09 & \textbf{0.30}$\pm$0.010 \\ \hline
\end{tabular}%
}
\end{table}
\section{Conclusion}
    In this paper, we address the problem of regret minimisation
using a bounded number of words of memory. This problem becomes interesting where the number
of arms is too large to consider all of them simultaneously, for example, crowd-sourcing,
drug testing \etc.
Some existing approaches~\citep{Herschkorn+PR:1996,Berry+CZHS:1997}
considers only the infinite bandit instances consist of Bernoulli arms. 
Recently, \citet{Liau+PSY:2018} present
an explore-then-commit based algorithm for finite bandit instances, which escapes such assumptions,
but very inefficient in practice.

We provide a \UCB~\citep{Auer+CF:2002} based algorithm \UCBM for finite bandit
instances, which is empirically far more efficient and
enjoys a sub-linear upper bound on the cumulative regret, but uses a bounded number of words of memory. Also, unlike all the existing algorithms, \UCBM offers the flexibility of varying the arm memory size, facilitating the experimenter to use the available memory resource. 
Further, we extend the existing algorithm  \textsc{QRM2}~\citep{RoyChaudhuri+K:2017}
for quantile-regret minimisation to  \QRMUCBM to achieve sub-linear quantile regret
under the bounded arm memory constraint. 
We empirically verify that \QRMUCBM incurs a lower conventional cumulative regret on
a various infinite bandit instances than the existing
algorithms~\citep{Herschkorn+PR:1996,Berry+CZHS:1997}, which needs $O(1)$ memory.

We find that providing a lower bound on the cumulative regret under the bounded arm memory constraint is an interesting question, and we leave that for future investigation.
\pagebreak
\bibliographystyle{icml2019}
\bibliography{rpbibfile}

\begin{thebibliography}{27}
\providecommand{\natexlab}[1]{#1}
\providecommand{\url}[1]{\texttt{#1}}
\expandafter\ifx\csname urlstyle\endcsname\relax
  \providecommand{\doi}[1]{doi: #1}\else
  \providecommand{\doi}{doi: \begingroup \urlstyle{rm}\Url}\fi

\bibitem[Agrawal(1995)]{Agrawal:1995}
Agrawal, R.
\newblock The continuum-armed bandit problem.
\newblock \emph{SIAM J. Control Optim.}, 33\penalty0 (6):\penalty0 1926--1951,
  1995.

\bibitem[Agrawal \& Goyal(2012)Agrawal and Goyal]{Agrawal+G:2012}
Agrawal, S. and Goyal, N.
\newblock Analysis of {T}hompson sampling for the multi-armed bandit problem.
\newblock In \emph{Proc. of the 25th Annual Conf. on Learning Theory},
  volume~23, pp.\  39.1--39.26, Edinburgh, Scotland, 2012. PMLR.

\bibitem[Agrawal \& Goyal(2013)Agrawal and Goyal]{Agrawal+G:2013}
Agrawal, S. and Goyal, N.
\newblock Further optimal regret bounds for thompson sampling.
\newblock In \emph{Proc. AISTATS 2013}, volume~31, pp.\  99--107. PMLR, 2013.

\bibitem[Aho et~al.(1974)Aho, Hopcroft, and Ullman]{Aho+HU:1974}
Aho, A.~V., Hopcroft, J.~E., and Ullman, J.~D.
\newblock \emph{The Design and Analysis of Computer Algorithms}.
\newblock Addison-Wesley, 1974.

\bibitem[Armitage(1960)]{Armitage:1960}
Armitage, P.
\newblock \emph{Sequential Medical Trials.}
\newblock Blackwell Scientific Publications, 1960.

\bibitem[Audibert \& Bubeck(2009)Audibert and Bubeck]{Audibert+B:2009}
Audibert, J.-Y. and Bubeck, S.
\newblock {Minimax policies for adversarial and stochastic bandits}.
\newblock In \emph{Proc. {COLT} 2009}, pp.\  217--226, 2009.

\bibitem[Auer et~al.(2002)Auer, Cesa-Bianchi, and Fischer]{Auer+CF:2002}
Auer, P., Cesa-Bianchi, N., and Fischer, P.
\newblock Finite-time analysis of the multiarmed bandit problem.
\newblock \emph{Machine Learning}, 47\penalty0 (2-3):\penalty0 235--256, 2002.

\bibitem[Berry \& Fristedt(1985)Berry and Fristedt]{Berry+F:1985}
Berry, D. and Fristedt, B.
\newblock \emph{Bandit Problems: Sequential Allocation of Experiments}.
\newblock Chapman \& Hall, 1985.

\bibitem[Berry et~al.(1997)Berry, Chen, Zame, Heath, and
  Shepp]{Berry+CZHS:1997}
Berry, D.~A., Chen, R.~W., Zame, A., Heath, D.~C., and Shepp, L.~A.
\newblock Bandit problems with infinitely many arms.
\newblock \emph{The Annals of Statistics}, 25\penalty0 (5):\penalty0
  2103--2116, 1997.

\bibitem[Bubeck et~al.(2009)Bubeck, Munos, and Stoltz]{Bubeck+MS:2009}
Bubeck, S., Munos, R., and Stoltz, G.
\newblock Pure exploration in multi-armed bandits problems.
\newblock In \emph{Algorithmic Learning Theory}, pp.\  23--37. Springer Berlin
  Heidelberg, 2009.

\bibitem[Chapelle \& Li(2011)Chapelle and Li]{Chapelle+L:2011}
Chapelle, O. and Li, L.
\newblock An empirical evaluation of thompson sampling.
\newblock In \emph{Advances in neural information processing systems}, pp.\
  2249--2257, 2011.

\bibitem[Colton(1963)]{Colton:1963}
Colton, T.
\newblock A model for selecting one of two medical treatments.
\newblock \emph{Journal of the American Statistical Association}, 58\penalty0
  (302):\penalty0 388--400, 1963.

\bibitem[Cormen et~al.(2009)Cormen, Leiserson, Rivest, and
  Stein]{Cormen+LRS:2009}
Cormen, T.~H., Leiserson, C.~E., Rivest, R.~L., and Stein, C.
\newblock \emph{Introduction to Algorithms, Third Edition}.
\newblock The MIT Press, 2009.

\bibitem[Cover(1968)]{Cover:1968}
Cover, T.~M.
\newblock A note on the two-armed bandit problem with finite memory.
\newblock \emph{Information and Control}, 12\penalty0 (5):\penalty0 371 -- 377,
  1968.

\bibitem[Cover et~al.(1976)Cover, Freedman, and Hellman]{Cover:1976}
Cover, T.~M., Freedman, M.~A., and Hellman, M.~E.
\newblock Optimal finite memory learning algorithms for the finite sample
  problem.
\newblock \emph{Information and Control}, 30\penalty0 (1):\penalty0 49 -- 85,
  1976.

\bibitem[Herschkorn et~al.(1996)Herschkorn, Peköz, and
  Ross]{Herschkorn+PR:1996}
Herschkorn, S.~J., Peköz, E., and Ross, S.~M.
\newblock Policies without memory for the infinite-armed {B}ernoulli bandit
  under the average-reward criterion.
\newblock \emph{Prob. in the Engg. and Info. Sc.}, 10\penalty0 (1):\penalty0
  21--28, 1996.

\bibitem[Isbell(1959)]{Isbell:1959}
Isbell, J.~R.
\newblock On a problem of robbins.
\newblock \emph{Ann. Math. Stat.}, 30\penalty0 (2):\penalty0 606--610, 06 1959.

\bibitem[Jamieson et~al.(2014)Jamieson, Malloy, Nowak, and
  Bubeck]{Jamieson+MNB:2014}
Jamieson, K.~G., Malloy, M., Nowak, R., and Bubeck, S.
\newblock lil' {UCB} : {A}n optimal exploration algorithm for multi-armed
  bandits.
\newblock In \emph{Proc. {COLT} 2014}, pp.\  423--439, 2014.

\bibitem[Kleinberg(2005)]{Kleinberg:2005}
Kleinberg, R.
\newblock Nearly tight bounds for the continuum-armed bandit problem.
\newblock In \emph{Adv. NIPS 17}, pp.\  697--704. MIT Press, 2005.

\bibitem[Liau et~al.(2018)Liau, Price, Song, and Yang]{Liau+PSY:2018}
Liau, D., Price, E., Song, Z., and Yang, G.
\newblock Stochastic multi-armed bandits in constant space.
\newblock In \emph{Proc. AISTATS 2018}, volume~84, pp.\  386--394. PMLR, 2018.

\bibitem[Pek\"{o}z(2003)]{Pekoz:2003}
Pek\"{o}z, E.~A.
\newblock Some memoryless bandit policies.
\newblock \emph{Journal of Applied Probability}, 40\penalty0 (1):\penalty0
  250--256, 2003.

\bibitem[Robbins(1952)]{Robbins:1952}
Robbins, H.
\newblock Some aspects of the sequential design of experiments.
\newblock \emph{Bulletin of the AMS}, 58\penalty0 (5):\penalty0 527--535, 1952.

\bibitem[Robbins(1956)]{Robbins:1956}
Robbins, H.
\newblock A sequential decision problem with a finite memory.
\newblock \emph{PNAS}, 42\penalty0 (12):\penalty0 920--923, 1956.

\bibitem[{Roy Chaudhuri} \& Kalyanakrishnan(2017){Roy Chaudhuri} and
  Kalyanakrishnan]{RoyChaudhuri+K:2017}
{Roy Chaudhuri}, A. and Kalyanakrishnan, S.
\newblock {PAC} identification of a bandit arm relative to a reward quantile.
\newblock In \emph{Proc. {AAAI 2017}}, pp.\  1977--1985. AAAI Press, 2017.

\bibitem[Roy~Chaudhuri \& Kalyanakrishnan(2018)Roy~Chaudhuri and
  Kalyanakrishnan]{RoyChaudhuri+K:2018}
Roy~Chaudhuri, A. and Kalyanakrishnan, S.
\newblock Quantile-regret minimisation in infinitely many-armed bandits.
\newblock In \emph{Proc. UAI 2018}, 2018.
\newblock To appear.

\bibitem[Tran-Thanh et~al.(2014)Tran-Thanh, Stein, Rogers, and
  Jennings]{Tran-Thanh+SRJ:2014}
Tran-Thanh, L., Stein, S., Rogers, A., and Jennings, N.~R.
\newblock Efficient crowdsourcing of unknown experts using bounded multi-armed
  bandits.
\newblock \emph{Artif. Intl.}, 214:\penalty0 89 -- 111, 2014.

\bibitem[Wang et~al.(2008)Wang, Audibert, and Munos]{Wang+AM:2008}
Wang, Y., Audibert, J.-Y., and Munos, R.
\newblock Algorithms for infinitely many-armed bandits.
\newblock In \emph{Adv. {NIPS 21}}, pp.\  1729--1736. Curran Associates Inc.,
  2008.

\end{thebibliography}
\pagebreak
\begin{appendices}
\clearpage
\section{Proofs from Section~\ref{subsec:ucbm}}
\label{app:algfinite}
 \begin{lemma}
   \label{lem:numsubphase}
   For a given $K$-sized set of arms $\A$, and an arm memory size $M < K$,
   the number of sub-phases required to ensure that each arm in
   $\A$ has been chosen into arm memory at least once is not more than
   $h_0$.
  \end{lemma}
  
  \begin{proof}
   We notice that at the beginning of each sub-phase there are exactly $M-1$
   arms except the arm $\hat{a}$ recommended from the previous step. 
   Let, $h$ be the maximum number of sub-phases possible in a phase.
   We realise that each phase $w$ ends as soon as for every arm $a \in \A$,
   there exists a sub-phase $j$, such that $S^{w,j} \ni a$.
   Therefore, $h = \min \{y : \A \subseteq \cup_{j=1}^y S^{w,j}\} = \bceil{\frac{K-1}{M-1}} = h_0$.
  \end{proof}

  \begin{lemma}
   \label{lem:maxnumphase}
   For a given $K$-sized set of arms $\A$, and an arm memory size $M < K$,
   the number of phases \UCBM executes is upper bounded by  
   $x_0 \defeq \bceil{\log\frac{2T}{MK}}$.
  \end{lemma}

  \begin{proof}
   Let $x$ be the total number of phases executed by \UCBM .
   It should be noted that the value of $M$, $K$, and $T$ might be such that
   the total horizon ($T$) runs out before finishing the last phase.
   Now, for any given phase $w$ ($w \geq 1$), the horizon spent on each
   sub-phase is the same, that is $b_w = 2^{w-1}b_1$. Therefore, we can write
   \begin{align*}
    &T = \sum_{w=1}^{x} \sum_{j=1}^{h_0} b_w = h_0 b_1\sum_{w=1}^x 2^{w-1} = h_0 b_1 (2^x -1),\\
    & \implies x = \log\left(\frac{T}{h_0 b_1}+1\right),\\
    & \leq \log\left(\frac{T}{\bceil{\frac{K-1}{M-1}}b_1}+1\right),\; \left[\text{because}, h_0=\bceil{\frac{K-1}{M-1}}\right],\\
    & \leq \log\left(\frac{T}{\bceil{\frac{K-1}{M-1}} M(M+2)}+1\right),\;\\
    & [\text{because}, b_1 =M(M+2)],\\
    & \leq \log\left(\frac{T}{\frac{K-1}{M-1} M(M+2)}+1\right) < \log\left(\frac{T}{K M}+1\right),\\
    & \leq \log\left(\frac{2T}{K M}\right)\; [\text{because}, T > 2MK ],\\
    & \implies x \leq \bceil{\log\frac{2T}{MK}} = x_0.
   \end{align*}
  \end{proof}
  
  \lemubsimpreg*
    \begin{proof}
    Letting, $\E[r_*^{y,j}] \defeq \E[\mu^* - \mu_*^{y,j}]$.
    We break the proof into two steps. 
    Step 1 upper bounds $\E[r_*^{y,h_0}]$, which is an upper bound
    on $\E[r_*^{y,j}]$, for all $j \in [h_0]$ ; while Step 2 upper bounds $\E[r_*^{y+1,j}]$. Both the
    steps are based on Corollary~\ref{cor:maxnsubphase}, that ensures at least
    one of the $h_0$ consecutive sub-phases (not necessarily from the same phase)
    must contain the optimal arm $a^*$ in the arm-memory.
    %
    
    \paragraph{Step 1.} Let, $1 \leq k_0 \leq h_0-1$ be the first sub-phase
    in phase $y$, to have $a^*$ in the arm-memory. Therefore, 
    $k_0 \defeq \min\{i \in [h_0]: a^* \in S^{y,i}\}$,
    and hence, by definition, $\E[r_*^{y, k_0+1}] = \E[r^{y, k_0+1}]$.
    Therefore, for any subsequent sub-phase $j \in \{k_0+1, \cdots, h_0\}$ in phase $y$,
    $\E[r_*^{y,j}] = \E[\mu^* - \mu_*^{y,j}] = \E[\mu^* -  \mu_*^{y,k_0+1}] + \sum_{v = k_0+2}^{j-1} \E[\mu_*^{y,v} - \mu_*^{y,v+1}]$.
    As there are $h_0$ sub-phases in any phase, hence, for all $k_0 + 1 \leq j \leq h_0$,
    $\E[r_*^{y,j}] \leq \E[r_*^{y,h_0}] \leq (h_0-k_0+1) \E[r^y] \leq h_0 \E[r^y]$.
    

    
    \paragraph{Step 2.} Let, $j_0$ be a sub-phase in phase $y-1$, such that $a^* \in S^{y-1, j_0}$.
    From Step 1,  
    $\E[r_*^{y-1,h_0}]  \leq (h_0- j_0 + 1) \E[r^{y-1}]$.
    Now, considering sub-phase $i$ in phase $y$, we realise that if $i \geq j_0$, then there exists a
    sub-phase $w \in \{1,\cdots,i\}$ such that $a^* \in S^{y, w}$. Now, for $i \leq j_0-1$,
    \begin{align*}
        & E[r_*^{y,i}] \leq \max_{2\leq j_0 \leq [h_0]} \max_{1\leq i \leq j_0-1} \E[r_*^{y-1,h_0}] + i \cdot \E[r^y],\\
        & \leq \max_{2\leq j_0 [h_0]} \max_{1\leq i \leq j_0-1} \leq (h_0- j_0 + 1) \E[r^{y-1}] + i\cdot \E[r^y],\\
        & \leq (h_0 - 1)  \E[r^{y-1}] +  \E[r^y] < 2 h_0 \E[r^y] \\
        &\hspace{3cm}[\text{Because} \E[r^{y}] < \E[r^{y-1}] < 2 \E[r^{y}]].
    \end{align*}
    Together, Step 1 and Step 2 prove the lemma.
  \end{proof}
  \lemsumrswj*
  \begin{proof}
   \begin{align*}
    & \sum_{w=1}^{x_0}\sum_{j=1}^{h_0} R_{w,j}^{(1)} = \sum_{j=1}^{h_0} R_{1,j}^{(1)} + \sum_{w=2}^{x_0}\sum_{j=1}^{h_0} R_{w,j}^{(1)}\\
    & \leq h_0 b_1 + \sum_{w=2}^{x_0}\sum_{j=1}^{h_0} R_{w,j}^{(1)}\\
    & \leq h_0 b_1 + \sum_{w=2}^{x_0}\sum_{j=1}^{h_0} b_w \E[\mu^* - \mu_*^{w,j}]\\
    & \leq h_0 b_1 + \sum_{w=2}^{x_0}\sum_{j=1}^{h_0} b_w (2h_0 \E[r^{w}] )\; [\text{using Lemma~\ref{lem:ubsimpreg}}]\\
    & \leq h_0 b_1 + 2 C_1 h_0^2 \sum_{w=2}^{x_0} b_w \sqrt{\frac{M\log b_w}{b_w}}\; [\text{using Corollary~\ref{cor:succsimpreg}}]\\
    & \leq h_0 b_1 + 2 C_1 h_0^2 \sum_{w=2}^{x_0} \sqrt{b_w M \log b_w}\\
    & \leq h_0 b_1 + 2 C_2 h_0^2\sqrt{M b_1} \sum_{w=2}^{x_0} \left(2^{w-1}\log\left(2^{w-1}b_1\right)\right)^\frac{1}{2}\\
    & \hspace{5cm} [\text{because}, b_w = 2^{w-1}b_1]\\
    & = h_0 b_1 + 2 C_2 h_0^2\sqrt{M b_1} \sum_{w=2}^{x_0} \left(\left(w-1 +\log b_1\right) 2^{w-1}\right)^\frac{1}{2}\\
    & \leq h_0 b_1 + C_3 h_0^2\sqrt{M b_1} \sum_{w=2}^{x_0} \left((w-1)2^{w-1}\right)^\frac{1}{2}\\
    &  \left[\text{because}, T > KM^2(M+2), \text{and}\; b_1 = M(M+2),\right.\\
    &  \hspace{0.5cm}\left. \text{therefore}, x_0-1 \geq \log b_1 = \log M(M+2)\right]\\
    & \leq h_0 b_1 + C_4 h_0^2\sqrt{M b_1} \left(x_0\cdot2^{x_0}\right)^\frac{1}{2}\\
    & \leq C_5 \left(\bbceil{\frac{K-1}{M-1}} M (M+2) +\right.\\
    & \hspace{1cm}  \left.\bbceil{\frac{K-1}{M-1}}^2\sqrt{M^2 (M+2)} \left(\frac{T}{MK}\log\frac{T}{MK}\right)^\frac{1}{2}\right)\\
    &  \hspace{3cm} [\text{substituting for } b_1, h_0\; \text{and}\; x_0]\\
    & \leq C_6 \left(\frac{K}{M} M^2 + \left(\frac{K}{M}\right)^2\sqrt{M^3} \left(\frac{T}{MK}\log\frac{T}{MK}\right)^\frac{1}{2}\right)\\
    & \leq C_7 \left(KM + \frac{K^{3/2}}{M}\sqrt{T\log\frac{T}{MK}}\right)\\
    & \leq C_8 \left(KM  + \frac{K^{3/2}}{M}\sqrt{T\log\frac{T}{MK}}\right),
   \end{align*}
   wherein, $C_1, C_2,\cdots,C_8$ are appropriate constants.
  \end{proof}
  
  \lemsumrcwj*
  \begin{proof}
   We notice that at any sub-phase $j$ of any phase $w \geq 2$, 
   due to Lemma~\ref{lem:ucb1ub}, there exists a
   constant $C$, such that $R_{w,j}^{(2)} \leq C \sqrt{b_w M \log b_w}$.
   Therefore,
   \begin{align*}
    & \sum_{w=1}^{x_0}\sum_{j=1}^{h_0} R_{w,j}^{(2)} = \sum_{j=1}^{h_0} R_{1,j}^{(2)} + \sum_{w=2}^{x_0}\sum_{j=1}^{h_0} R_{w,j}^{(2)},\\ 
    & \leq h_0 b_1 + \sum_{w=1}^{x_0}\sum_{j=1}^{h_0} C \sqrt{b_w M\log b_w}\; [\text{using Lemma~\ref{lem:ucb1ub}}],\\
    & = h_0 b_1 +  h_0 C\sum_{w=1}^{x_0}\sqrt{2^{w-1}b_1 M \log \left(2^{w-1}b_1\right)},\\
    &  \hspace{3cm}  [\text{substituting for } b_w]\\
    & \leq \bbceil{\frac{K-1}{M-1}} M(M+2) +\\
    &  \hspace{1cm}  C_1 \bbceil{\frac{K-1}{M-1}} \sqrt{M^2(M+2)} \sum_{w=1}^{x_0}\left((w-1)2^{w-1}+\right.\\
    &  \hspace{1cm}\left.2^{w-1}\log (M(M+2))\right)\;[\text{substituting for}\; h_0\; \text{and}\; b_1],\\
    & \leq C_2 \left(\frac{K-1}{M-1} M(M+2) +\right.\\
    &  \hspace{1cm}\left. \frac{K-1}{M-1} \sqrt{M^2(M+2)} \sum_{w=1}^{x_0}\sqrt{(w-1)2^{w-1}}\right)\\
    &  \hspace{3cm} \left[\text{because}, T > KM^2 (M+2)\right]\\
    & \leq C_3 \left(\frac{K}{M} M^2 + \frac{K}{M} M^{3/2} \sqrt{x_0 2^{x_0}}\right)\\
    & = C_3 \left(KM + K\sqrt{M} \sqrt{\frac{T}{MK} \log \frac{T}{MK}}\right),\\
    & \leq C_4 \left(KM +  \sqrt{TK \log \frac{T}{MK}}\right),
   \end{align*}
   wherein, $C_1, C_2, C_3, C_4$ are appropriate constants.
  \end{proof}
\clearpage
\section{Additional Experimental Results from Section~\ref{subsec:exptfinite}}
\label{app:experiment_finite}

\begin{figure}[H]
 \centering
    \subfigure[]{\label{fig:app:k10reg}\includegraphics[width=0.8\columnwidth]{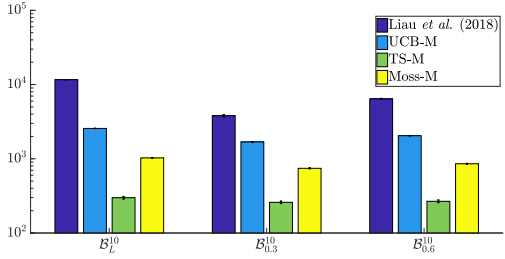}}
    \subfigure[]{\label{fig:app:k10pull}\includegraphics[width=0.8\columnwidth]{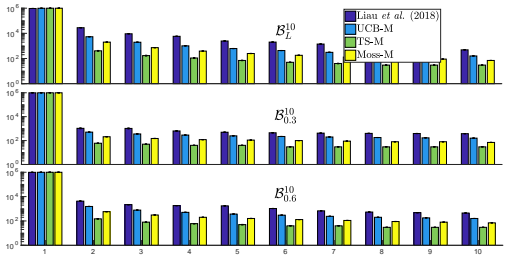}}
    \caption{Comparison of incurred cumulative regret (\ref{fig:app:k10reg}) 
	     and the number of pulls to the individual arms (\ref{fig:app:k10pull})
	     on the instances $\B_L^{10}$, $\B_{0.3}^{10}$ and $\B_{0.6}^{10}$
	     by algorithms \textsc{UCB-M}, \textsc{TS-M}, \textsc{Moss-M} for $M=2$, and the
	     algorithm of \cite{Liau+PSY:2018}, after $10^6$ pulls.
	     Each bar represents the 
	     average over 100 iterations, and
	     with one standard error. For details about the instances and the
	     algorithms we refer to Section~\ref{subsec:exptfinite}.
	     }
    \label{fig:app:k10}
\end{figure}


\begin{figure}[H]
 \centering
 \subfigure[]{\label{fig:varmem_B100A03_h106}\includegraphics[width=0.8\columnwidth]{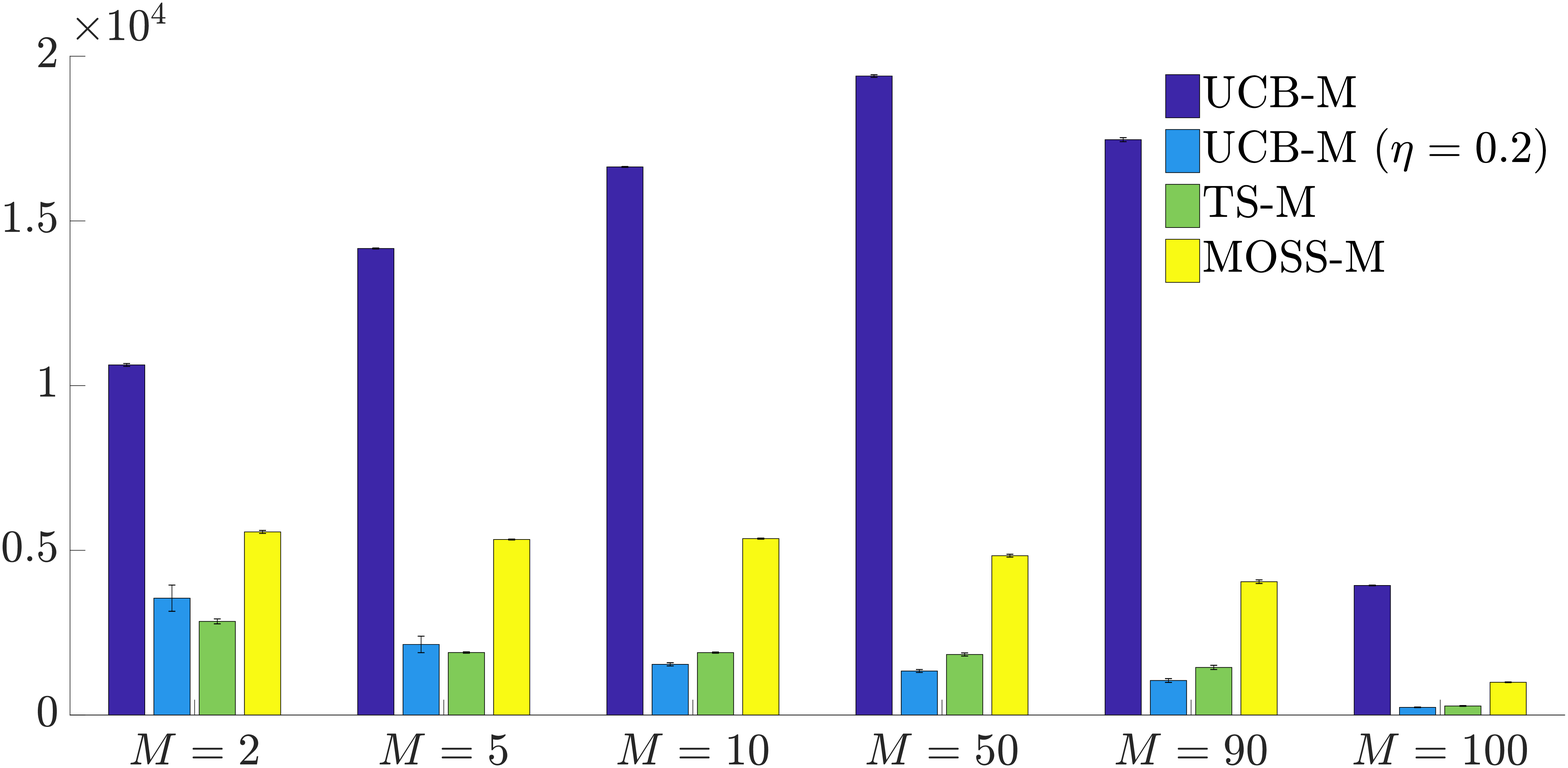}}
 \caption{Comparison of incurred regret on the instance $\B_{0.3}^{100}$. Each bar represents 
	  regret incurred after
	  $10^6$ pulls, averaged over 100 iterations, and
	     with one standard error. For details about the instances and the
	     algorithms we refer to Section~\ref{subsec:exptfinite}.}
 \label{fig:varmem_B100A03}
\end{figure}

\begin{figure}[H]
 \centering
 \subfigure[]{\label{fig:varmem_B100A06_h106}\includegraphics[width=0.8\columnwidth]{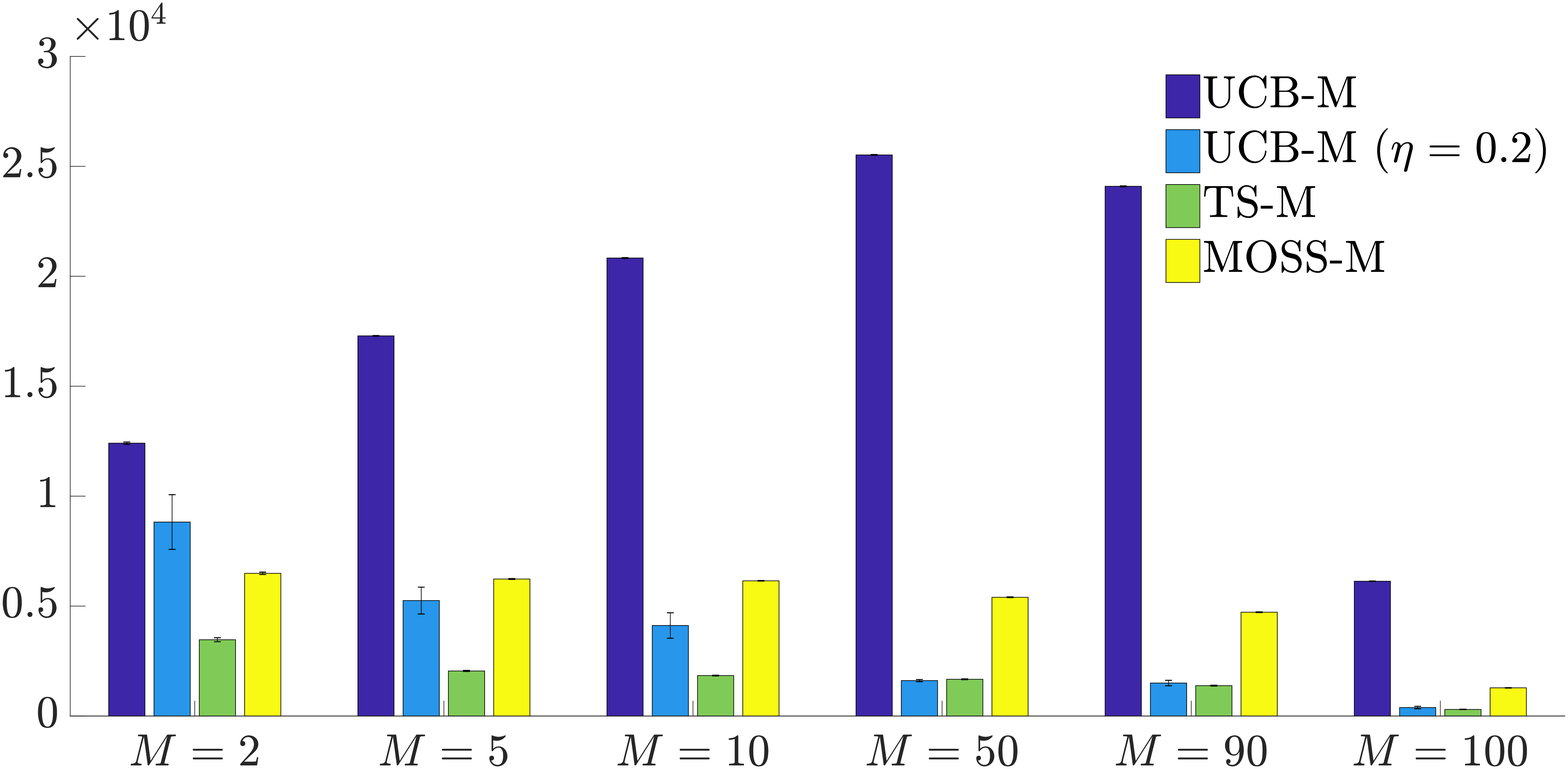}}
 \caption{Comparison of incurred regret on the instance $\B_{0.6}^{100}$. Each bar represents 
	  regret incurred after
	  $10^6$ pulls, averaged over 100 iterations, and
	     with one standard error. For details about the instances and the
	     algorithms we refer to Section~\ref{subsec:exptfinite}.}
 \label{fig:varmem_B100A06}
\end{figure}

\clearpage
\section{Proofs from Section~\ref{subsec:qrm2ucbmalg}}
\label{app:qrm2ucbm}
In this appendix we provide the materials to complete the proof of Theorem~\ref{thm:ubqrm2ucbm}.
\begin{restatable}{lemma}{lemrhostar}
\label{lem:rhostar}
Let, $r^* =  \bceil{\frac{1}{\alpha}\log\left(\frac{1}{\rho}\log\frac{1}{\rho}\right) - \log B}$. Then, for every phase $r \geq r^*$, the size of $\K_r$
can be lower bounded as $n_r = \bceil{t_r^\alpha} \geq \bceil{\frac{\alpha\log\mathrm{e}}{(1+\gamma)\rho}\cdot\ln t_r}$,
wherein, $0.53 < \gamma \defeq \max_{x} \frac{\log\log x}{\log x} < 0.531$.
\end{restatable}
\begin{proof}
We notice, for every, $r \geq r^*$, $t_r \geq \bceil{\left(\frac{1}{\rho}\log\frac{1}{\rho}\right)^\frac{1}{\alpha}}$.
Then, for each $r \geq r^*$, we can lower bound the size of the set $\K_r$ as follows. 
As, $|\K_r| = n_r = \bceil{t_r^\alpha}$ is an integer, to ease
the calculation let us define $s_u = 2^{u}B$, where $u \in \mathbb{R}^+$, and therefore, $s_u \in \mathbb{R}^+$ does not need to be an integer.
Now, letting $u^* \defeq \log\left(\frac{1}{\rho}\log\frac{1}{\rho}\right)^\frac{1}{\alpha}$, we get
\begin{align}
\label{eq:rhostar}
& \frac{1}{\rho}\log s_{u^*} \nonumber\\
& = \frac{1}{\rho}\log\left(\frac{1}{\rho}\log\frac{1}{\rho}\right)^\frac{1}{\alpha}, \nonumber\\
& = \frac{1}{\alpha\rho} \left(\log\frac{1}{\rho} + \log\log\frac{1}{\rho}\right), \nonumber\\
& = \frac{1}{\alpha\rho}\log\frac{1}{\rho} \left(1 + \frac{\log\log\frac{1}{\rho}}{\log\frac{1}{\rho}}\right), \nonumber\\
& \leq \frac{1+\gamma}{\alpha\rho}\log\frac{1}{\rho},\; \left[\text{as}\; \gamma \defeq \max_x \frac{\log\log x}{\log x}\right]\nonumber\\
& = \frac{1+\gamma}{\alpha}\left(\frac{1}{\rho}\log\frac{1}{\rho}\right), \nonumber\\
& = \frac{1+\gamma}{\alpha}s_{u^*}^\alpha. \nonumber\\
\implies & s_{u^*}^\alpha \geq \frac{\alpha}{(1+\gamma)\rho}\log s_{u^*} \nonumber\\
\implies &s_{u^*}^\alpha \geq  \frac{\alpha \log e}{(1+\gamma)\rho}\ln s_{u^*}.
\end{align}
As, $s_u^{\alpha}$ grows with $u$ faster than $\log s_u$, therefore,
\begin{align*}
\forall u \geq u^*,\;s_{u^*}^\alpha & \geq \frac{\alpha \log e}{(1+\gamma)\rho}\ln s_{u^*}
\end{align*}
Therefore, recalling that $r$ is an integer, for all values of $r \geq \ceil{r^*}$, the statement of the lemma follows.
\end{proof}

\begin{lemma}
\label{lem:qreger}
The expected regret due to not encountering any arm from the set
$\mathcal{TOP}_\rho$ is during running of the algorithm, is in $O\left(\left(\frac{1}{\rho}\log\frac{1}{\rho}\right)^\frac{1}{\alpha} +  T^{1-\frac{\alpha\log\mathrm{e}}{1+\gamma}}\right)$.
\end{lemma}
\begin{proof}

 We define an event that no arm from $\mathcal{TOP}_\rho$ is in $\K_r$ as $E_r(\rho) \defeq \{\K_r \cap \mathcal{TOP}_\rho = \emptyset\}$, and note $\Pr\{E_r(\rho)\} = (1-\rho)^{n_r}$. Now, for some  $\alpha \in (0,1)$ that shall be tuned later, let $r^* = \ceil{(1/\alpha)\log((1/\rho)\log(1/\rho))}$. 
 Therefore, in the round $r^*$, the number of pulls
 is given by $t_{r^*} = 2^{r^*} = ((1/\rho)\log(1/\rho))^{1/\alpha}$.
 Now, for $r \geq r^*$,  the number of arms in $\K_r$ is given by $n_r = t_r^\alpha \geq \ceil{(\alpha/((1+\gamma)\rho))\cdot\ln t_r^{\log\mathrm{e}}}$,
 wherein, $\gamma = \max_{x} (\log\log x) / \log x$ ($0.53 < \gamma < 0.531$). 

 Therefore,
 $\Pr\{E_r(\rho)\} $ 
 $= (1-\rho)^{n_r}$
 $\leq \exp(- \ceil{(\alpha/((1+\gamma)))\cdot\ln t_r^{\log\mathrm{e}}})$
 $\leq {t_r}^{-\alpha\log\mathrm{e}/(1+\gamma)}$.

Using Lemma~\ref{lem:rhostar}, 
below we present the detailed steps for obtaining \eqref{eq:regempty2} in the proof of Theorem~\ref{thm:ubqrm2ucbm}.
\begin{align*}
 & \sum_{r=1}^{\log (T/B)} t_r \Pr\{E_r(\rho)\}\\
 & = \sum_{r=1}^{r^*-1} t_r \Pr\{E_r(\rho)\}  + \sum_{r=r^*}^{\log (T/B)} t_r \Pr\{E_r(\rho)\}\\
 & \leq \sum_{r=1}^{r^*-1} t_r + \sum_{r=r^*}^{\log (T/B)} {t_r}^{1-\frac{\alpha\log\mathrm{e}}{1+\gamma}}\\
 &  \leq t_{r^*} + \sum_{r=r^*}^{\log  (T/B)} {t_r}^{1-\frac{\alpha\log\mathrm{e}}{1+\gamma}} \\
 &  = t_{r^*} + \sum_{r=r^*}^{\log  (T/B)} {(B2^r)}^{1-\frac{\alpha\log\mathrm{e}}{1+\gamma}} \\
 & \leq  t_{r^*} + B^{1-\frac{\alpha\log\mathrm{e}}{1+\gamma}}\sum_{j=1}^{\log (T/B)-r^*} \left(\frac{T}{B2^{j}} \right)^{1-\frac{\alpha\log\mathrm{e}}{1+\gamma}}\\
 & \leq B \cdot 2^{\bceil{\log\left(\frac{1}{\rho}\log\frac{1}{\rho}\right)^\frac{1}{\alpha}-\log B}} +  T^{1-\frac{\alpha\log\mathrm{e}}{1+\gamma}}\sum_{j=0}^{\log T-r^*} \left(\frac{1}{2}\right)^{j(1-\frac{\alpha\log\mathrm{e}}{1+\gamma})}\\
 & < 2^{{\log\left(\frac{1}{\rho}\log\frac{1}{\rho}\right)^\frac{1}{\alpha}}+1}  +  T^{1-\frac{\alpha\log\mathrm{e}}{1+\gamma}}\sum_{j=0}^{\infty} \left(\frac{1}{2}\right)^{j(1-\frac{\alpha\log\mathrm{e}}{1+\gamma})}\\
 & = O\left(2^{\log\left(\frac{1}{\rho}\log\frac{1}{\rho}\right)^\frac{1}{\alpha}}  +  T^{1-\frac{\alpha\log\mathrm{e}}{1+\gamma}}\right)\\
 & = O\left(\left(\frac{1}{\rho}\log\frac{1}{\rho}\right)^\frac{1}{\alpha} +  T^{1-\frac{\alpha\log\mathrm{e}}{1+\gamma}}\right)\\
 & = O\left(\left(\frac{1}{\rho}\log\frac{1}{\rho}\right)^\frac{1}{\alpha} +  T^{1-\frac{\alpha\log\mathrm{e}}{1+\gamma}}\right).
\end{align*}
\end{proof}

\begin{lemma}
\label{lem:qregne}
For $r^*$ defined in Lemma~\ref{lem:rhostar}, given that for all $r \geq r^*$, algorithm 
\QRMUCBM has encountered at least one arm from $\mathcal{TOP}_\rho$, the incurred regret beyond the round $r^*$ is not more than 
$C'  \left(M T^{\alpha} + \frac{\sqrt{\log M}}{M}\sqrt{T^{1+3\alpha}\log\frac{T}{M}}\right)$;
for some constant $C'$.
\end{lemma}

\begin{align*}
 &\sum_{r=r^*}^{\log (T/B)} C \left(n_r M + \frac{1}{M}\sqrt{n_r^3 t_r \log \frac{t_r}{n_r M}}\right), \\
 &\sum_{r=r^*}^{\log (T/B)} C \left(t_r^\alpha M + \frac{1}{M}\sqrt{ t_r^{1+3\alpha} \log \frac{t_r}{n_r M}}\right), \\
 & = \sum_{r=r^*}^{\log (T/B)} C \left(2^{\alpha r} B^{\alpha}M + \frac{1}{M}\sqrt{2^{(1+3\alpha)r}B^{(1+3\alpha)}\log \left(\frac{B}{M}  2^{(1-\alpha)r}\right)}\right), \\
 & = \sum_{r=r^*}^{\log (T/B)} C \left(2^{\alpha r}B^\alpha M + \frac{\sqrt{B^{(1+3\alpha)}\log(B/M)}}{M}\sqrt{(1-\alpha)r2^{(1+3\alpha)r}}\right), \\
 &\leq  \sum_{r=r^*}^{\log (T/B)} C_1 \left(2^{\alpha r}B^\alpha M + \frac{\sqrt{B^{(1+3\alpha)}\log(B/M)}}{M}\sqrt{r2^{(1+3\alpha)r}}\right) \\
&\leq  \sum_{r=r^*}^{\log (T/B)} C_1 \left(\left(\frac{T}{2^jB}\right)^{\alpha}B^\alpha M + \frac{\sqrt{B^{(1+3\alpha)}\log(B/M)}}{M}\sqrt{\left(\frac{T}{B2^j}\right)^{1+3\alpha}\log\frac{T}{B}}\right) \\
 & = \sum_{j=0}^{\log (T/B)-r^*} C_2 \left(M\left(\frac{T}{2^j}\right)^{\alpha} + \frac{\sqrt{\log M}}{M}\sqrt{\left(\frac{T}{2^j}\right)^{1+3\alpha}\log\frac{T}{B}}\right),\\
 & \leq C_3 \left(MT^{\alpha}\sum_{j=0}^{\log (T/B)-r^*}\left(\frac{1}{2^{\alpha}}\right)^{j} + \right.\\ 
 & \hspace{1.2cm}\left.\frac{\sqrt{\log M}}{M}T^{(1+3\alpha)/2}\sqrt{\log\frac{T}{B}} \sum_{j=0}^{\log (T/B)-r^*}\sqrt{\left(\frac{1}{2^j}\right)^{1+3\alpha}}\right),\\
 & \leq C_4 \left(M T^{\alpha} + \frac{\sqrt{\log M}}{M}\sqrt{T^{1+3\alpha}\log\frac{T}{B}}\right),\\
 & \leq C_5 \left(M T^{\alpha} + \frac{\sqrt{\log M}}{M}\sqrt{T^{1+3\alpha}\log\frac{T}{M}}\right) [\text{because}, B = \left(M^2(M+2)\right)^\frac{1}{1-\alpha}]
\end{align*}
for some constants $C_1, C_2, C_3, C_4$ and $C_5$.
\clearpage
\section{Additional Experimental Results from Section~\ref{subsec:exptinfinite}}
\label{app:experiment_infinite}
    

\begin{table}[H]
\caption{Cumulative regret ($/10^{5}$) of \protect\QRMUCBM, \protect\QRMTSM, \protect\QRMMSM and the strategies proposed by \cite{Herschkorn+PR:1996} and \cite{Berry+CZHS:1997} after $10^6$ pulls, on instances $I_1$, $I_2$, $I_3$ and $I_4$. Each result is the average of 20 runs, showing one standard error.}
\label{tab:compare_herschberry}
\resizebox{\columnwidth}{!}{%
\begin{tabular}{|l|c|l|l|l|l|}
\hline
\multicolumn{1}{|c|}{Algorithms} & M & \multicolumn{1}{c|}{\begin{tabular}[c]{@{}c@{}}$I_1$: $\beta(0.5,2)$\\ $\mu^*=1$\end{tabular}} & \multicolumn{1}{c|}{\begin{tabular}[c]{@{}c@{}}$I_2$: $\beta(1,1)$\\ $\mu^*=1$\end{tabular}} & \multicolumn{1}{c|}{\begin{tabular}[c]{@{}c@{}}$I_3$: $\beta(0.5,2)$\\ $\mu^*=0.6$\end{tabular}} & \multicolumn{1}{c|}{\begin{tabular}[c]{@{}c@{}}$I_4$: $\beta(1,1)$\\ $\mu^*=0.6$\end{tabular}} \\ \hline
\begin{tabular}[c]{@{}l@{}}Non-stationary Policy\\ \citep{Herschkorn+PR:1996}\end{tabular} & 1 & 3.58 $\pm$0.4 & 1.11 $\pm$0.2 & 1.64 $\pm$ 0.2 & 0.79 $\pm$ 0.1 \\ \hline
\begin{tabular}[c]{@{}l@{}}$\sqrt{T}$-run\\ \citep{Berry+CZHS:1997}\end{tabular} & 2 & 6.18$\pm$0.5 & 1.11$\pm$0.4 & 4.18$\pm$0.3 & 2.03$\pm$0.3 \\ \hline
\begin{tabular}[c]{@{}l@{}}$\sqrt{T}\ln T$-learning\\ \citep{Berry+CZHS:1997}\end{tabular} & 2 & 6.32$\pm$0.4 & 0.69$\pm$0.3 & 4.38$\pm$0.2 & 2.15$\pm$0.3 \\ \hline
\begin{tabular}[c]{@{}l@{}}Non-recalling $\sqrt{T}$-run\\ \citep{Berry+CZHS:1997}\end{tabular} & 1 & 5.35 $\pm$0.5 & 0.03 $\pm$0.004 & 4.56 $\pm$ 0.001 & 2.55 $\pm$ 0.001 \\ \hline
\rule{0pt}{4ex}
\multirow{2}{*}{\textsc{QUCB-M}} & $2$ & 3.69$\pm$0.34 & 0.74$\pm$0.11 & 2.27$\pm$0.21 & 0.51$\pm$0.07 \\ \cline{2-6}
\rule{0pt}{4ex} 
 & $10$ & 4.26$\pm$0.37 & 0.91$\pm$0.19 & 2.65$\pm$0.22 & 0.63$\pm$0.11 \\ \hline
 \rule{0pt}{4ex}
\multirow{2}{*}{\textsc{QUCB-M} $\eta=0.2$} & $2$ & 3.67$\pm$0.35 & 0.72$\pm$0.12 & 2.21$\pm$0.21 & 0.55$\pm$0.08 \\ \cline{2-6} 
\rule{0pt}{4ex}
& $10$ & 4.15$\pm$0.36 & 0.79$\pm$0.19 & 2.51$\pm$0.22 & 0.54$\pm$0.11 \\ \hline
\rule{0pt}{4ex}
\multirow{2}{*}{\textsc{QTS-M}} & $2$ & 3.14$\pm$0.39 & 0.62$\pm$0.07 & 1.97$\pm$0.19 & 0.44$\pm$0.07 \\ \cline{2-6}
\rule{0pt}{4ex} 
 & $10$ & 3.88$\pm$0.35 & 0.67$\pm$0.13 & 2.49$\pm$0.23 & 0.45$\pm$0.06 \\ \hline
 \rule{0pt}{4ex}
\multirow{2}{*}{\textsc{QMoss-M}} & $2$ & 3.64$\pm$0.34 & 0.70$\pm$0.11 & 2.21$\pm$0.21 & 0.46$\pm$0.07 \\ \cline{2-6} 
\rule{0pt}{4ex}
 & $10$ & 4.16$\pm$0.36 & 0.80$\pm$0.19 & 2.53$\pm$0.22 & 0.52$\pm$0.11 \\ \hline
\end{tabular}%
}
\end{table}

For $\alpha=0.205$ the algorithms explore very small number of arms, that causes
incorporating a good arm very unlikely leading to a high regret.

\end{appendices}
\end{document}